\documentclass[11pt]{article}
\usepackage{amsmath,amssymb,amsthm}
\usepackage{graphicx}      
\usepackage{booktabs}
\usepackage{hyperref}
\usepackage{fullpage}
\usepackage{float}         
\usepackage{listings}      
\usepackage{xcolor}        
\usepackage{caption}
\usepackage{subcaption}
\usepackage{placeins}      

\usepackage{setspace}
\usepackage[authoryear,round]{natbib}

\newtheorem{Proposition}{Proposition}[section]

\newtheorem{Theorem}{Theorem}[section]
\newtheorem{Lemma}{Lemma}[section]

\theoremstyle{definition}

\theoremstyle{remark}
\newtheorem*{Remark}{Remark}

\theoremstyle{plain}
\newtheorem*{TheoremMain}{Theorem~\ref*{thm:gpdhp-stability}}
\newtheorem*{PropositionMain}{Proposition~\ref*{prop:growth}}

\title{A Semiparametric Discrete Hawkes Model with a Collapsed Gaussian-Process Prior}
\author{
Trinnhallen Brisley \\
University of Edinburgh \\
\texttt{t.brisley@sms.ed.ac.uk}
\and
Gordon Ross \\
University of Edinburgh \\
\texttt{gordon.ross@ed.ac.uk}
\and
Daniel Paulin \\
Nanyang Technological University \\
\texttt{daniel.paulin@ntu.edu.sg}
}

\newcommand{\ThmGeoErgBody}{%
Let the counts follow
\(N(t)\mid\mathcal{H}(t-1)\sim\mathrm{NB}\{\lambda(t),\kappa\}\) with
\(\lambda(t)=h_s\bigl\{b+\textstyle\sum_{d=1}^{D}f(d)\,N(t-d)\bigr\}\),
where \(D:=D_{\max}<\infty\), the excitation kernel \(f\in\mathbb{R}^{D}\) is fixed,
\(h_s(x)=s\log(1+e^{x/s})+\epsilon\) with link scale \(s>0\) and floor \(\epsilon\ge0\),
\(\kappa\in(0,\infty)\), and the baseline is constant, \(b(t)\equiv b\in\mathbb{R}\).
Assume
\[
R_+\;:=\;\sum_{d=1}^{D}f^{+}_d\;<\;1,\qquad f^{+}_d:=\max\{f(d),0\}.
\]
Then the window chain \(X(t):=\bigl(N(t),\dots,N(t-D+1)\bigr)\) is an irreducible, aperiodic,
positive Harris recurrent Markov chain on \(\mathbb{N}_0^{D}\) with a unique stationary
distribution \(\pi\), and:
\begin{enumerate}
\item[(i)] (Geometric ergodicity.) There exist strictly positive weights
\(\omega_1,\dots,\omega_D\), depending only on \(f\), such that, with
\(V(x):=1+\sum_{d=1}^{D}\omega_dx_d\), we have \(\pi(V)<\infty\) and, for some
\(M<\infty\) and \(\bar\varrho\in(0,1)\),
\[
\bigl\|P^{n}(x,\cdot)-\pi\bigr\|_{V}\;\le\;M\,\bar\varrho^{\,n}\,V(x)
\qquad\text{for all }x\in\mathbb{N}_0^{D},\ n\ge0,
\]
where \(\|\mu\|_{V}:=\sup_{|g|\le V}|\mu(g)|\).
\item[(ii)] (Stationary mean.) \(\mathbb{E}_\pi\{N(t)\}\le\bigl(b_++s\log2+\epsilon\bigr)/(1-R_+)\),
with \(b_+:=\max\{b,0\}\).
\end{enumerate}%
}

\newcommand{\PropGrowthBody}{%
Let the counts follow
\(N(t)\mid\mathcal{H}(t-1)\sim\mathrm{NB}\{\lambda(t),\kappa\}\) with
\(\lambda(t)=h_s\bigl\{b(t)+\textstyle\sum_{d=1}^{D}f(d)\,N(t-d)\bigr\}\),
with \(D\), \(f\), \(h_s\), and \(\kappa\) as in Theorem~\ref*{thm:gpdhp-stability} but with an
arbitrary deterministic baseline sequence \(b(\cdot)\), finite at each \(t\), and fix any
initial distribution for the initial window \(X(0)\in\mathbb{N}_0^{D}\).
\begin{enumerate}
\item[(i)] (Existence.) There is a process \((N(t))_{t\ge1}\), unique in law, with exactly these
conditional distributions and initial law, and it satisfies
\(\Pr\{N(t)<\infty\text{ for all }t\ge1\}=1\). This holds for every kernel
\(f\in\mathbb{R}^{D}\) and every such baseline, with no condition on \(R_+\).
\item[(ii)] (Input-tracking growth.) If \(R_+<1\) and the initial window has finite mean,
\(\mathbb{E}\bigl\{\sum_{d=1}^{D}X_d(0)\bigr\}<\infty\), then there exist \(\varrho\in(0,1)\)
and \(C_f<\infty\), depending only on \(f\), such that for every \(t\ge1\),
\[
\mathbb{E}\{N(t)\}\;\le\;C_f\,\varrho^{\,t}\,
\mathbb{E}\Bigl\{1+\sum_{d=1}^{D}X_d(0)\Bigr\}
\;+\;1\;+\;\frac{s\log2+\epsilon+\max_{1\le r\le t}b(r)_+}{1-\varrho},
\, b(r)_+:=\max\{b(r),0\}.
\]
In particular \(\mathbb{E}\{N(t)\}\) is uniformly bounded in \(t\) whenever
\(\sup_{t\ge1}b(t)<\infty\), and in general grows at most in proportion to the running maximum
\(\max_{r\le t}b(r)_+\) (at most linearly in \(t\) for the periodic-plus-linear-trend baseline).
\end{enumerate}%
}

\begin{document}

\maketitle

\begin{abstract}
Hawkes processes are used in settings where past events increase the likelihood of future events occurring, resulting in a natural clustering structure. Traditional Hawkes process models treat events as occurring in continuous time, but in many applications only the number of events occurring within a sequence of time bins is observed. We propose the Gaussian Process Discrete Hawkes Process (GP-DHP), a semiparametric model for discrete-time self-exciting count data that places Gaussian-process priors on both the baseline and the excitation. Marginalizing the two GP components induces a single latent Gaussian trajectory. A finite-rank factorization of this collapsed prior permits maximum a posteriori (MAP) estimation without forming or factorizing a \(T\times T\) covariance matrix. External covariates can enter the baseline through the same construction. In simulations, GP-DHP recovers diverse excitation shapes and evolving baselines. In applications to weekly disease surveillance (Singapore dengue and German cryptosporidiosis), daily shooting counts (New York City and the Gun Violence Archive), and a daily worldwide terrorism-incident series, it attains the best held-out predictive accuracy on four of the five series and, on the fifth, an accuracy not significantly different from the best.
\end{abstract}

\section{Introduction}

Hawkes processes are self-exciting stochastic models in which the probability of an event occurring increases in response to past occurrences. Since their introduction by \citet{hawkes1971}, they have been widely adopted across disciplines such as seismology \citep{ogata1988}, finance \citep{bacry2015}, criminology \citep{mohler2011}, social networks \citep{zhou2013, du2016}, and epidemiology \citep{reinhart2018, browning2021}. Their appeal lies in the ability to model bursty, temporally clustered behavior via an additive decomposition of the event rate into baseline and excitation components.

Much of the literature has focused on continuous-time Hawkes processes, where the conditional intensity evolves over continuous time and is typically expressed as the sum of a deterministic baseline and a parametric excitation kernel. Common excitation functions include exponential and power-law decay due to their simplicity and interpretability \citep{ogata1988}. More recent work proposes flexible alternatives using histogram-based kernels \citep{lewis2011}, basis expansions \citep{zhou2013}, neural network parameterizations \citep{mei2017}, and Gaussian-process priors on the triggering kernel itself \citep{zhang2020, zhou2020, malemshinitski2021}, with fully Bayesian approaches for latent network inference \citep{linderman2014}. A related line places Gaussian-process priors on the intensity of an inhomogeneous \emph{Poisson} (Cox) process, which captures a smoothly varying rate but no self-excitation \citep{adams2009, lloyd2015, samo2015}.

In many real-world scenarios, however, events are recorded at fixed intervals. Examples include weekly case counts in public health and daily incident logs in security contexts. Applying continuous-time Hawkes models to such data requires time discretization, which can introduce bias and hinder interpretability. Discrete Hawkes processes (DHPs), closely related to integer-valued autoregressive processes \citep{kirchner2016hawkes}, provide a principled alternative: they define event intensity directly over discrete time steps while retaining the usual self-exciting structure. Recent work has also studied more general nonlinear discrete-time Hawkes models, including stability questions for processes with inhibition \citep{costa2024general,costa2024stability}. Despite these advances, most existing DHP models rely on restrictive parametric assumptions. The baseline intensity is often modeled as constant or sinusoidal, while excitation kernels are restricted to geometric or negative-binomial forms. While effective in some settings, such assumptions can limit the ability to capture long memory, nonstationarity, or changes in excitation over time.

To address these limitations, \citet{browning2022} introduced a nonparametric DHP using a random histogram prior over the excitation kernel with trans-dimensional MCMC for inference. While this approach allows data-driven excitation structure, the use of a fixed intercept for the baseline may limit its ability to capture smooth or evolving background dynamics. Moreover, the sampling scheme can be costly for long processes. 

In this paper, we take a different approach and introduce the \textit{Gaussian Process Discrete Hawkes Process (GP-DHP)}, a semiparametric model for discrete-time self-exciting count data. Because many regularly observed event series exhibit known calendar periodicities, we model the baseline using a finite-Fourier seasonal GP component, giving a parsimonious prior over periodic background variation without fixing a deterministic seasonal curve. For the excitation, we combine a standard negative-binomial Hawkes kernel with a Gaussian-process correction. The parametric component supplies a standard Hawkes memory kernel, while the GP component allows departures from that kernel when supported by the data. Since the excitation remains linear in the latent GP component, the additive intensity has a collapsed Gaussian representation, enabling MAP inference and an interpretable decomposition into exogenous (baseline) and endogenous (excitation) components while avoiding direct optimization over separate high-dimensional latent functions.

The remainder of the paper is organized as follows. Section~\ref{sec:dhp-background} reviews the discrete Hawkes background, Section~\ref{sec:model} introduces the GP-DHP model and its baseline, excitation, and covariate components, and Section~\ref{Inference} develops the collapsed-latent inference, the forward-validation hyperparameter selection, and the component projection. Section~\ref{sec:experiments} reports the simulation and real-data experiments, and Section~\ref{sec:conclusion} concludes.

\section{Discrete Hawkes Background}\label{sec:dhp-background}

\subsection{Discrete Hawkes Process}\label{univariate-discrete-hawkes}

The discrete-time Hawkes process (DHP) is the discrete analogue of the continuous-time Hawkes process introduced by \citet{hawkes1971}. In continuous time, the intensity function defines the instantaneous rate of event occurrence and typically increases immediately following a self-exciting or clustering phenomenon. In contrast, the DHP operates over discrete time steps, \(t \in \mathbb{N}\), and models the expected number of events per interval using a combination of exogenous and endogenous contributions.

\begin{figure}
  \centering
  \includegraphics[width=0.75\linewidth]{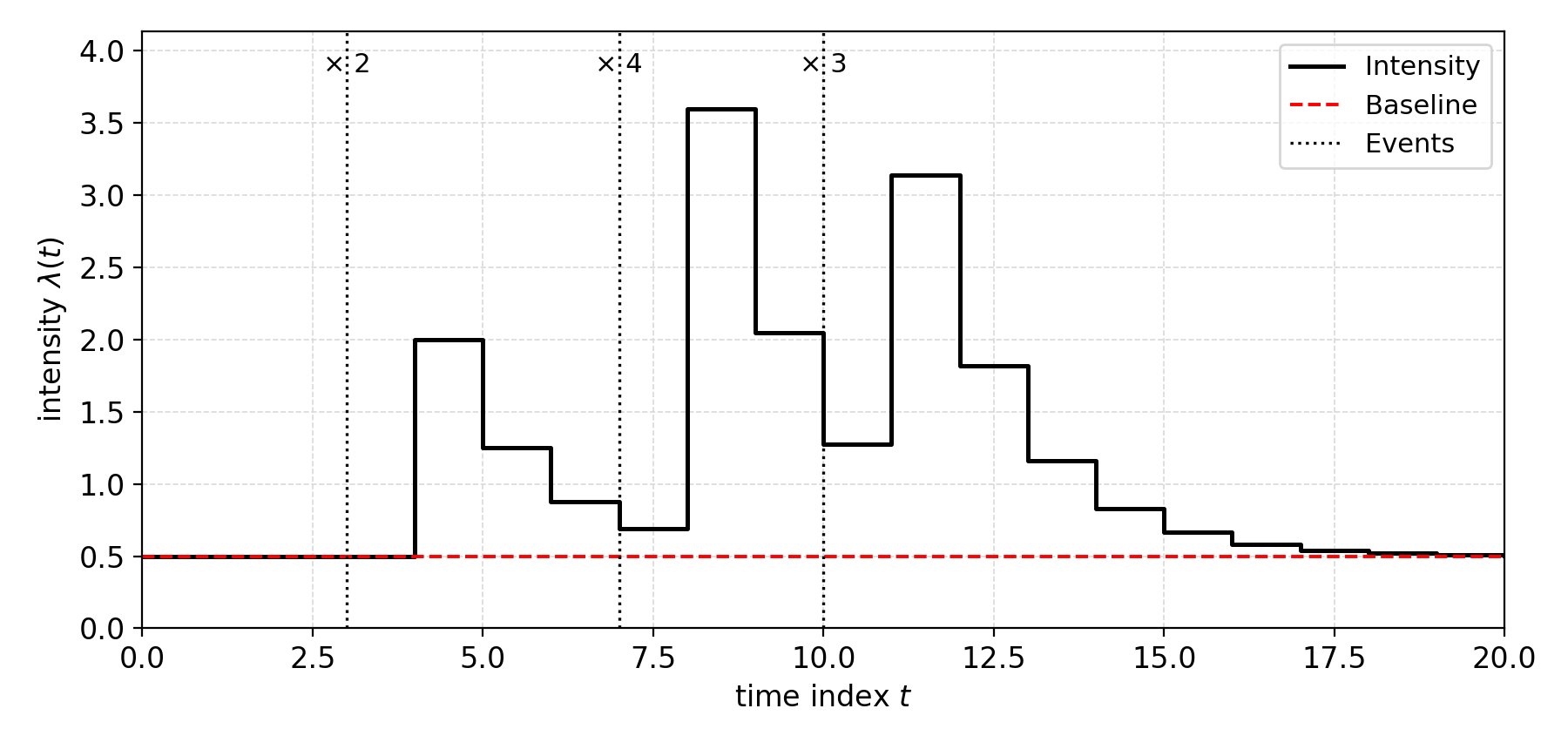}
  \caption{Discrete-time Hawkes intensity on \([0,20]\) with geometric excitation. The plot shows intensity (black step), baseline (red dashed), and event times (black dotted; multiplicities annotated at the top). Parameters: baseline \(\mu=0.5\), jump size 0.75, and geometric decay parameter 0.5; events at \(t=3,7,10\) with multiplicities \(2,4,3\), respectively.}
  \label{fig:discrete-hawkes-toplabels}
\end{figure}

The DHP is designed for count data observed at regular time intervals, such as hourly, daily, or weekly series. It captures the intuition that recent events tend to increase the probability of future events, a property known as self-excitation. This makes DHPs particularly useful for modeling clustered sequences, such as infectious disease cases, financial transactions, or crime incidents, where new occurrences can trigger follow-up events in subsequent intervals. Figure~\ref{fig:discrete-hawkes-toplabels} gives a simple illustration of the resulting discrete-time intensity path.

Let \(N(t) \in \mathbb{N}_0\) denote the number of events observed during the interval \((t-1,t]\), and let the event history up to time \(t-1\) be denoted \(\mathcal{H}(t-1)=\{N(s):s<t\}\). The conditional intensity function, or expected event rate at time \(t\), is defined as
\begin{equation}\label{eq:discrete-hawkes}
    \lambda(t)=\mathbb{E}\{N(t)\mid \mathcal{H}(t-1)\}
    =\mu(t)+\sum_{d=1}^{t-1}N(t-d)\Phi(d),
\end{equation}
where:
\begin{itemize}
    \item \(\mu(t)>0\) is the baseline intensity, representing spontaneous (i.e. exogenous) events not triggered by past observations. It can capture systematic background variation beyond the influence of past events. In practice, this often reflects smooth trends or seasonal cycles. For example, in epidemiology \(\mu(t)\) may encode annual patterns in disease incidence, while in security contexts it may capture differences between weekdays and weekends.
    \item \(\Phi(d)\) is the excitation kernel, describing the influence that events occurring \(d\) steps in the past exert on the present rate. In the standard linear Hawkes model this kernel is nonnegative, and stability is controlled by the total branching ratio \(R=\sum_{d\geq 1}\Phi(d)\). 
\end{itemize}

The observed counts could then be modeled using a Poisson distribution conditional on the past:
\[
N(t)\mid\mathcal{H}(t-1) \sim \mathrm{Poisson}\{\lambda(t)\}.
\]

However in many real-world data sets, observations may be overdispersed relative to the Poisson model. As such, in the model below we instead use a negative-binomial observation layer, with \(\lambda(t)\) as its conditional mean. This separates the modelling of temporal self-excitation from the modelling of residual count variation around the intensity; the Poisson model is obtained as a limiting case.


\vspace{1em}
\subsubsection{Branching Representation of the Discrete Hawkes Process}
The discrete Hawkes model above is often interpreted through the Hawkes cluster representation: events arise either as \emph{immigrants} from the baseline \(\mu(t)\) or as \emph{offspring} triggered by past events, with integer lags distributed according to \(\Phi(d)\) \citep{hawkes1971, hawkes1974}. In this view, bursts form around immigrant arrivals as offspring generate further descendants. We call this the branching process interpretation.

This branching process interpretation can facilitate practical inference and simulation. Each event at time \(t\) is assigned a parent source, either an earlier event or a spontaneous immigrant. The set of all such parent--child relationships is known as the branching structure.

Let \(N(t)\in\mathbb{N}_0\) denote the observed number of events at time \(t\), and define a latent vector
\[
\boldsymbol{n}^{(t)}=(n_0^{(t)},n_1^{(t)},\ldots,n_{t-1}^{(t)}),
\]
where \(n_0^{(t)}\) is the number of immigrant events at time \(t\), and \(n_d^{(t)}\) is the number of offspring events at time \(t\) triggered by events at time \(t-d\), for \(d=1,\ldots,t-1\). These quantities satisfy
\[
N(t)=\sum_{d=0}^{t-1}n_d^{(t)}.
\]
Given the branching structure \(\{\boldsymbol{n}^{(t)}\}_{t=1}^T\), the complete-data likelihood factorizes into contributions from baseline and excitation sources. The probability of an event being triggered by a parent at time \(t-d\) is proportional to \(N(t-d)\Phi(d)\), while the probability of being an immigrant is proportional to \(\mu(t)\). Thus the branching assignments can be viewed as draws from
\[
(n_0^{(t)},n_1^{(t)},\ldots,n_{t-1}^{(t)})
\sim
\operatorname{Multinomial}\left(
N(t);\frac{\mu(t)}{\lambda(t)},
\frac{N(t-1)\Phi(1)}{\lambda(t)},\ldots,
\frac{N(1)\Phi(t-1)}{\lambda(t)}
\right).
\]

The branching representation is exact for the standard linear Poisson DHP in Equation~\eqref{eq:discrete-hawkes}. The semiparametric model below uses a rectified link and a negative-binomial observation layer, so we use this representation as background and as motivation for Hawkes-style excitation summaries rather than as a literal offspring-count construction for inference.

Branching-based Gibbs samplers are useful for many parametric Hawkes models, but they are less attractive here. With Gaussian-process priors on both the baseline and excitation, the branching allocations and the latent functions are strongly coupled: changing the allocation of events can imply global changes in the fitted baseline or excitation surface. In preliminary experiments this led to slow mixing and expensive high-dimensional updates. We therefore use a collapsed latent-intensity MAP formulation, described in Section~\ref{Inference}, rather than sampling the branching variables explicitly.

\section{The GP-DHP Model}\label{Proposed Model}\label{subsec:GP-DHP}\label{sec:model}

We now define the semiparametric GP-DHP model. The term semiparametric reflects that the conditional count distribution and the negative-binomial excitation backbone are specified parametrically, while the baseline and the flexible lag component of the excitation are assigned Gaussian-process priors. A Gaussian process is a prior over functions under which any finite collection of evaluations is jointly Gaussian, specified by a mean and a covariance (kernel) function that encodes structure such as smoothness, periodicity, or trends \citep{rasmussen2006gaussian}. 

We model the latent additive intensity \(\ell\) as a discrete Hawkes process (as in Equation~\eqref{eq:discrete-hawkes}). That is, the sum of two components: a latent baseline contribution \( b(t) \)---on the pre-link scale, and unlike the classical nonnegative baseline intensity \(\mu(t)\) of Section~\ref{sec:dhp-background} it may be negative---and a time-lagged excitation term parameterized by a function \( f(d) \),
\begin{equation}\label{eq:lt-decomp}
    \ell(t)=b(t)+\sum_{d=1}^{D_{\max}}N(t-d)f(d),
\end{equation}
where \(N(t)\) denotes the observed event count in interval \((t-1,t]\), and \(d\in\{1,2,\dots\}\) indexes discrete lags. To obtain a nonnegative conditional mean (the discrete-time analogue of an intensity, which we call the intensity throughout), we apply a smooth rectifying link to the latent process:
\[
\lambda(t)=h_s\{\ell(t)\},\qquad h_s(x)=s\,\log\{1+\exp(x/s)\}+\epsilon,
\]
a softplus link with scale \(s>0\) and a small fixed numerical floor \(\epsilon>0\) (not a modelling hyperparameter). As \(s\to0\), \(h_s\) tends to the hard rectified (ReLU-type) link \(\max\{x,0\}+\epsilon\), so the resulting model is a nonlinear discrete-time Hawkes model with a rectified link, in the sense of recent work on discrete-time Hawkes processes with inhibition and signed effects \citep{costa2024general}. A positive \(s\) removes the non-differentiability at the origin and makes both the MAP optimization and the gradient-based hyperparameter search below better behaved; we select \(s\) jointly with the other hyperparameters (Section~\ref{sec:hyperparameter-selection}). Because the link acts pointwise on \(\ell(t)\), the collapsed-prior and projection results in Section~\ref{Inference} do not depend on this particular choice of link.

Next, we use a negative-binomial likelihood for the observed counts,
\begin{equation}\label{eq:nb-observation}
\begin{gathered}
N(t)\mid\mathcal{H}(t-1) \sim \mathrm{NB}\{\lambda(t),\kappa\},\\
\mathbb{E}\{N(t)\mid\mathcal{H}(t-1)\}=\lambda(t),
\qquad
\operatorname{Var}\{N(t)\mid\mathcal{H}(t-1)\}=\lambda(t)+\tfrac{\lambda(t)^2}{\kappa}.
\end{gathered}
\end{equation}
This observation model is useful for binned event series whose conditional variability is larger than that allowed by a Poisson likelihood. The Hawkes structure specifies the conditional intensity \(\lambda(t)\), while the size parameter \(\kappa\) controls residual count dispersion around that intensity. As \(\kappa\to\infty\), the likelihood reduces to the Poisson likelihood.

As in any additive Hawkes decomposition, \(b(t)\) and the excitation contribution \(\sum_d N(t-d)f(d)\) are not uniquely identifiable without additional structure. Slowly varying temporal patterns can otherwise be attributed either to the baseline or to accumulated excitation. We address this by using a seasonal baseline prior for calendar-scale variation and an excitation prior that combines a standard parametric memory kernel with a flexible GP correction. The projection in Section~\ref{sec:decomposition} then gives a unique minimum-prior-norm decomposition of the optimized latent trajectory.

\paragraph{GP prior for the baseline \(b(t)\).}
Many regularly observed event series have known calendar periodicities, such as annual variation in weekly disease counts or seasonal patterns in daily incident records. We encode such structure through a finite-Fourier seasonal prior. Let
\[
\boldsymbol{h}_K(t)=\left(\sin\frac{2\pi t}{P},\cos\frac{2\pi t}{P},\ldots,
\sin\frac{2\pi Kt}{P},\cos\frac{2\pi Kt}{P}\right)^\top,
\]
and let \(H_K\) be the matrix with rows \(\boldsymbol{h}_K(t)^\top\).  We use
\begin{equation}\label{eq:Kb-fourier}
    K_b = \sigma_{\mathrm{level}}^2\mathbf{1}\mathbf{1}^\top
    + \sigma_{\mathrm{lin}}^2\mathbf{t}\mathbf{t}^\top
    + \sigma_{\mathrm{season}}^2 H_KH_K^\top
    {}+ \sigma_{\mathrm{week}}^2 H_{K_w}^{(P_w)}H_{K_w}^{(P_w)\top},
\end{equation}
where \(\mathbf{1}\) is the vector of ones, \(\mathbf{t}=(1,\ldots,T)^\top\), \(P\) is the known seasonal period, and \(K\) is the number of included harmonics. Setting \(\sigma_{\mathrm{lin}}=0\) omits the linear trend component. This construction is equivalent to placing Gaussian priors on the level, trend, and harmonic coefficients and integrating them out, while allowing separate prior scales for the constant level, the trend, and the seasonal variation. It is therefore a stochastic seasonal GP prior, not an empirical seasonal average or a fixed deterministic seasonal curve. The structured terms define a \emph{finite-rank} Gaussian prior supported on the span of the level, trend, and harmonic bases---a degenerate Gaussian on \(\mathbb{R}^T\), equivalently \(\boldsymbol{b}=B\theta_b\) with \(\theta_b\sim\mathcal N(0,I)\) (Section~\ref{sec:map-objective}); Equation~\eqref{eq:Kb-fourier} is this probabilistic covariance \(K_b\). Where a numerical factorization requires a strictly positive-definite matrix we use \(K_b^{\mathrm{num}}=K_b+\epsilon_b^2 I\) with a negligible fixed \(\epsilon_b\); this nugget is not part of the probabilistic prior and plays no role in the MAP objective or the component projection of Section~\ref{sec:decomposition}.

The fourth term is a \emph{weekly} harmonic block: \(H_{K_w}^{(P_w)}\) has the same finite-Fourier form as \(H_K\) but at the short period \(P_w=7\) with \(K_w\) harmonics, and \(\sigma_{\mathrm{week}}\) is its prior scale. Daily event series routinely carry a strong day-of-week cycle, and without a weekly baseline term the deterministic part of that cycle can only be represented through the lagged-count excitation, where it appears as a period-7 oscillation in the fitted lag response. Including the weekly block lets the baseline absorb the deterministic day-of-week level directly; as discussed in Section~\ref{sec:components}, any period-7 structure that remains in the fitted excitation is then attributable to lag-seven dependence beyond the fitted weekly baseline, rather than to baseline leakage. For the weekly-aggregated disease series we set \(\sigma_{\mathrm{week}}=0\) (no within-week structure to model); for the daily series \(\sigma_{\mathrm{week}}\) is selected by forward validation like the other scales, and the selected values are positive (Supplementary Table~\ref*{tab:supp-selected-hyperparameters}).

\paragraph{GP prior for the excitation kernel \(f(d)\).}
The parametric component of the excitation prior is a negative-binomial kernel on the finite lag window. Let
\begin{equation}\label{eq:m-nb}
    q_{\mathrm{NB}}(d;r,p_{\mathrm{lag}})
    =
    \frac{\tilde q_{\mathrm{NB}}(d-1;r,p_{\mathrm{lag}})}
    {\sum_{j=1}^{D_{\max}} \tilde q_{\mathrm{NB}}(j-1;r,p_{\mathrm{lag}})},
    \qquad d=1,\ldots,D_{\max},
\end{equation}
where \(\tilde q_{\mathrm{NB}}\) is the negative-binomial probability mass function on nonnegative integers. The excitation kernel is
\begin{equation}\label{eq:bridge-prior}
    f(d)=K_{\mathrm{NB}}q_{\mathrm{NB}}(d;r,p_{\mathrm{lag}})+u(d),
    \qquad u\sim\mathcal{GP}(0,K_g),
    \qquad K_{\mathrm{NB}}\geq0 .
\end{equation}
Here \(K_{\mathrm{NB}}\) controls the mass of the parametric kernel, while \(u(d)\) is a GP correction over lags.

A common way to construct nonnegative GP-based intensities is to transform a latent GP through a nonlinear link. Such constructions are natural in GP point-process models, but a nonlinear link on the excitation would destroy the Gaussian collapse used here, since the lagged excitation contribution would no longer be linear in the GP. The additive form above keeps the excitation linear in the GP component while allowing departures from the negative-binomial memory kernel.

We use the following lag-dependent covariance for \(u(d)\). Let
\[
a(d)=\sigma_g\exp\left(-\frac{\beta d}{2}\right),
\qquad
 w(d)=\frac{1-e^{-\beta d}}{\beta\ell_g},
\]
with \(\sigma_g\geq0\) an amplitude scale, \(\ell_g>0\) a base length-scale, and \(\beta\geq 0\) controlling lag attenuation. The limiting value \(\sigma_g=0\) removes the GP correction. Define
\[
k_{\mathrm{RBF}}\{w(d),w(d')\}
=
\exp\left[-\frac{1}{2}\{w(d)-w(d')\}^2\right].
\]
The covariance of the GP lag component is then
\begin{equation}\label{eq:Kf}
K_g(d,d') = a(d)a(d') k_{\mathrm{RBF}}\{w(d),w(d')\}+\epsilon_g^2\delta_{dd'}.
\end{equation}
This construction yields smooth, short-lag correlations while simultaneously shrinking long-lag variability via \(a(d)\) and compressing large lags through the warp \(w(\cdot)\). As \(\beta \to 0\), we recover a stationary RBF prior on the original lag scale because \(a(d)\!\to\! \sigma_g\) and \(w(d)\!\to\! d/\ell_g\). For \(\beta>0\), the prior increasingly attenuates and decorrelates remote lags, which discourages the GP component from absorbing slow-moving trends that should be attributed to the baseline process \(b(t)\).

Supplementary Figure~\ref*{fig:gp-excitation-beta-sweep} illustrates how the GP lag prior behaves as the attenuation parameter \(\beta\) increases. The top row shows sample draws of \(u(d)\) under different \(\beta\) values, highlighting how larger \(\beta\) suppresses long-lag fluctuations and concentrates variation near the origin. The bottom row displays the corresponding covariance heatmaps \(K_g\).

\paragraph{External covariates.}
When external covariates are available, they enter the model through the baseline, keeping the collapsed Gaussian structure intact. Given covariate series \(z_1(t),\ldots,z_J(t)\), we augment the latent intensity as
\begin{equation}\label{eq:cov-baseline}
    \ell(t)=b(t)+\sum_{j=1}^{J}c_j\,z_j(t)+\sum_{d=1}^{D_{\max}}N(t-d)f(d),
    \qquad c_j\sim\mathcal{N}(0,\sigma_{\mathrm{cov},j}^2),
\end{equation}
with the coefficients \(c_j\) given independent Gaussian priors and integrated out, exactly as for the baseline harmonics. This adds \(\sum_j \sigma_{\mathrm{cov},j}^2\, z_j z_j^\top\) to \(K_b\), so the covariates are simply extra columns of the collapsed baseline design and do not change the inference or projection machinery. Because the effect is linear in the latent intensity and passes through the same rectifying link, a covariate can raise or lower the conditional rate but cannot by itself make it negative. Each prior scale \(\sigma_{\mathrm{cov},j}\) is chosen by forward validation, with related covariates grouped so that they share a scale (for instance, one scale for a temperature driver and one shared scale for a block of holiday indicators); a selected scale at the bottom of its search box effectively removes a covariate. In the daily shooting applications we use a deseasonalized daily temperature anomaly and a set of per-holiday indicator series, described in Section~\ref{sec:real-data}.

\paragraph{Discussion.}
The finite-Fourier baseline prior and the excitation prior provide complementary inductive biases. Calendar-scale background variation is represented in \(b(t)\), while the excitation explains short- to medium-range triggering. The negative-binomial kernel gives a standard Hawkes memory backbone, and the GP component allows data-adaptive departures from that backbone. This prior structure is important because the two components enter additively in~\eqref{eq:lt-decomp}; without it, slowly varying seasonal behavior can be confounded with accumulated excitation.

\section{Inference}\label{Inference}

Our inference procedure centers on the latent intensity trajectory
\(\boldsymbol{\ell}=[\ell(1),\ldots,\ell(T)]^\top\), defined in Equation~\eqref{eq:lt-decomp}. Conditional on the negative-binomial excitation kernel and the GP hyperparameters, this trajectory has a Gaussian prior with a known mean and covariance. Rather than estimating the baseline and excitation functions directly during optimization, we work with this collapsed prior over \(\boldsymbol{\ell}\). MAP estimation is then performed on a single latent trajectory, and the fitted trajectory is subsequently projected back into baseline and excitation components.

While discrete Hawkes models are often interpreted using a branching structure, we do not use this representation for inference. As discussed in Section~\ref{sec:dhp-background}, branching-based Gibbs updates are natural for parametric DHPs, but become poorly conditioned when the branching variables are coupled to high-dimensional GP-distributed baseline and excitation functions. The collapsed latent-intensity representation avoids explicit branching variables and gives a deterministic optimization problem.

\subsection{Collapsed GP Prior}

Inference over separate baseline and excitation functions is difficult because the excitation enters through lagged sums of the observed count process. The additive latent trajectory itself has a Gaussian prior after centering by the parametric part of the excitation kernel. This allows us to optimize over the latent intensity trajectory rather than over the two component functions directly.

Let \(X\in\mathbb{R}^{T\times D_{\max}}\) be the lagged-count design matrix
\[
X_{t,d}=
\begin{cases}
N(t-d), & 1\leq d\leq D_{\max}\text{ and }d<t,\\
0, & \text{otherwise},
\end{cases}
\qquad t=1,\ldots,T.
\]
Let \(m_{\mathrm{NB}}\in\mathbb{R}^{D_{\max}}\) be the vector with entries \(K_{\mathrm{NB}}q_{\mathrm{NB}}(d;r,p_{\mathrm{lag}})\). Then
\[
\boldsymbol{\ell}=X m_{\mathrm{NB}}+\boldsymbol{z},
\qquad
\boldsymbol{z}=\boldsymbol{b}+X\boldsymbol{u},
\]
with \(\boldsymbol{b}\sim\mathcal{N}(0,K_b)\), \(\boldsymbol{u}\sim\mathcal{N}(0,K_g)\), and \(\boldsymbol{b}\) independent of \(\boldsymbol{u}\). Hence \(\boldsymbol{\ell}\) has the collapsed, possibly degenerate Gaussian law
\begin{equation}\label{eq:collapsed-bridge-prior}
\boldsymbol{\ell}\sim
\mathcal{N}\left(Xm_{\mathrm{NB}},\;K_z\right),
\qquad
K_z=K_b+XK_gX^\top.
\end{equation}

The law~\eqref{eq:collapsed-bridge-prior} is the probabilistic model, stated in the collapsed trajectory space. For computation, however, we work in an equivalent whitened, finite-dimensional parameterization. The baseline is the finite basis expansion \(\boldsymbol{b}=B\theta_b\), whose design columns \(B\in\mathbb{R}^{T\times d_b}\) are the scaled level, linear-trend, annual-harmonic and (for daily series) weekly-harmonic terms of Equation~\eqref{eq:Kb-fourier}, with \(\theta_b\sim\mathcal{N}(0,I)\); the GP correction is \(\boldsymbol{u}=L\theta_g\), where \(L\) is a Cholesky factor of the \(D_{\max}\times D_{\max}\) lag covariance (\(LL^\top=K_g\)) and \(\theta_g\sim\mathcal{N}(0,I)\). Stacking \(\theta=[\theta_b^\top,\theta_g^\top]^\top\) gives
\begin{equation}\label{eq:whitened}
\boldsymbol{\ell}=Xm_{\mathrm{NB}}+A\theta,\qquad A=[\,B \mid XL\,],\qquad \theta\sim\mathcal{N}(0,I),
\end{equation}
Equation~\eqref{eq:whitened} is the finite-rank prior: the centered trajectory \(\boldsymbol{\ell}-Xm_{\mathrm{NB}}=A\theta\) is supported on the \((d_b+d_g)\)-dimensional column space of \(A\), a degenerate Gaussian on \(\mathbb{R}^T\) with covariance \(K_z=AA^\top\). If a numerical nugget \(\epsilon_b^2 I\) is added to \(K_b\), the covariance corresponding to~\eqref{eq:collapsed-bridge-prior} becomes full rank; this nugget-augmented \(K_b^{\mathrm{num}}\) is not the fitted probabilistic prior, and is excluded from the MAP objective and the projection, so no independent per-time baseline residual is fitted. This parameterization avoids all \(T\times T\) linear algebra. The whitened coefficient \(\theta=[\theta_b^\top,\theta_g^\top]^\top\in\mathbb{R}^{d_b+d_g}\) is low-dimensional, \(d_b\) is a handful of baseline columns and \(d_g\le D_{\max}\), so all inference is carried out over \(\theta\): there is no \(T\)-dimensional covariance to invert. This is a change of coordinates, not a departure from the collapsed formulation: \(\theta\) parameterizes the same collapsed law~\eqref{eq:collapsed-bridge-prior}, and the uncollapsed alternative, joint optimization over the component functions themselves, is the comparator of Section~\ref{sec:collapsed-vs-uncollapsed}. The MAP fit (Section~\ref{sec:map-objective}) minimizes a smooth ridge-penalized negative log-likelihood in \(\theta\), and the projection (Section~\ref{sec:decomposition}) reads \(b(t)\) and \(f(d)\) off \(\theta\) directly.

\subsection{MAP Objective}\label{sec:map-objective}

Once the finite-dimensional representation of Equation~\eqref{eq:whitened} is available, inference reduces to optimization over the coefficient vector \(\theta\). Let \(N=[N(1),\ldots,N(T)]^\top\) denote the observed count vector; for fixed hyperparameters, the MAP coefficient \(\theta^*\) maximizes
\begin{equation}\label{eq:map-theta}
\log p(N\mid Xm_{\mathrm{NB}}+A\theta,\kappa)-\tfrac{1}{2}\theta^\top\theta,
\end{equation}
the conditional count log-likelihood penalized by the unit ridge from \(\theta\sim\mathcal N(0,I)\), and the fitted trajectory is \(\boldsymbol{\ell}^*=Xm_{\mathrm{NB}}+A\theta^*\). On the support \(\boldsymbol{\ell}\in Xm_{\mathrm{NB}}+\operatorname{col}(A)\) this ridge is the penalty of the degenerate finite-rank prior, written through the Moore--Penrose pseudo-inverse of \(K_z=AA^\top\) (Appendix~A); no \(T\times T\) system is formed. The intensity is \(\lambda(t)=h_s\{\ell(t)\}\), the softplus link of Section~\ref{sec:model}. Under the negative-binomial observation model in Equation~\eqref{eq:nb-observation}, the log-likelihood is
\[
\log p(N\mid\boldsymbol{\ell},\kappa)
=
\sum_{t=1}^T
\log \mathrm{NB}\{N(t);\lambda(t),\kappa\},
\]
where the negative-binomial distribution is parameterized by mean \(\lambda(t)\) and size \(\kappa\). Reported predictive log-likelihoods include all normalizing constants.

\subsection{Stability of the Fitted Process}\label{sec:stability}

Before selecting hyperparameters we record the stability property that motivates a constraint we impose throughout. The fitted one-step conditional mean is \(\lambda(t)=h_s\{b(t)+\sum_{d\ge1} f(d)\,N(t-d)\}\), a rectified-linear feedback in the past counts through the excitation coefficients \(f(d)\), which the Gaussian-process correction may render negative (inhibitory). The link is \(1\)-Lipschitz and tends to the floored hard rectifier \((\cdot)_++\epsilon\) as \(s\to0\); in that limit with \(\epsilon=0\), a constant baseline, and Poisson counts, the model is the discrete-time Hawkes process with inhibition of \citet{costa2024general}, whose Theorem~2.1 makes
\[
R_+ \;=\; \sum_{d=1}^{D_{\max}}\max\{\widehat f(d),\,0\}\;<\;1
\]
sufficient for geometric ergodicity: only the positive, excitatory part of the kernel enters the threshold, so inhibition relaxes it. In Appendix~B we prove the corresponding guarantees for the model actually fitted here, for any dispersion \(\kappa\in(0,\infty)\) and link scale \(s>0\). Existence is not at issue: the discrete-time process exists at all finite times with probability one for \emph{every} kernel and baseline (Proposition~\ref{prop:growth}). What \(R_+<1\) buys are the guarantees: under a constant baseline, geometric ergodicity of the count process (Theorem~\ref{thm:gpdhp-stability}); under an arbitrary deterministic baseline (seasonal, trended, or covariate-driven, conditionally on its path), mean growth at most of the order of the running maximum of the positive part of the baseline, hence uniformly bounded means whenever that path is bounded above (Proposition~\ref{prop:growth}). The condition is sufficient but not necessary: for memory length two, \citet{costa2024stability} show the exact stability region strictly exceeds \(\{R_+<1\}\) under inhibition. Rather than report \(R_+\) as a diagnostic a fit may or may not satisfy, we \emph{impose} it: every GP-DHP fit in Section~\ref{sec:experiments} is selected under the constraint \(R_+\le1-\delta\) with \(\delta=10^{-4}\), so every fitted kernel provably satisfies \(R_+<1\).

We state the two guarantees in full here; the proofs, together with the supporting lemmas and sharpness remarks (in particular, \(R_+<1\) cannot be weakened to \(R_+\le1\)), are in Appendix~B.

\begin{Theorem}[geometric ergodicity]\label{thm:gpdhp-stability}
\ThmGeoErgBody
\end{Theorem}

\begin{Proposition}[existence and input-tracking growth]\label{prop:growth}
\PropGrowthBody
\end{Proposition}

\subsection{Hyperparameter Selection}\label{sec:hyperparameter-selection}

Let \(\psi\) collect the model hyperparameters: the baseline and covariate scales, the negative-binomial excitation parameters, the GP lag-kernel parameters, the softplus scale \(s\), and the observation size \(\kappa\). We select \(\psi\) by forward validation of the one-step predictive log-likelihood. The fitting period is split chronologically into an inner training segment and a later validation segment. For each candidate \(\psi\), the MAP coefficient \(\widehat\theta_\psi\) is fitted on the training segment and scored on the validation segment,
\begin{equation}\label{eq:fv-criterion}
\mathrm{FV}(\psi)
=
\sum_{t\in\mathcal{V}}
\log p\{N(t)\mid\mathcal{H}(t-1);\psi,\widehat\theta_\psi\}.
\end{equation}
Each validation prediction uses the observed history available at that time. Candidates for which \(K_g\) cannot be stably factorized, or for which the MAP optimization fails to converge, are assigned a score of \(-\infty\).

This gives a bilevel optimization problem. The inner problem,
\[
\widehat\theta_\psi
=
\arg\min_\theta F_{\mathrm{tr}}(\theta;\psi),
\]
is the collapsed MAP fit on the inner training segment, while the outer objective \(G(\psi)=-\mathrm{FV}(\psi)\) evaluates its predictive performance. We differentiate the outer objective through the fitted inner solution using the implicit-function theorem. In the whitened coordinates of Equation~\eqref{eq:whitened}, the resulting adjoint hypergradient is
\begin{equation}\label{eq:fv-hypergrad}
\nabla_\psi G
=
\partial_\psi G
-
\big(\partial_\psi\nabla_\theta F_{\mathrm{tr}}\big)^\top\boldsymbol{\nu},
\qquad
\boldsymbol{\nu}
=
\big[\nabla^2_{\theta\theta}F_{\mathrm{tr}}\big]^{-1}
\partial_\theta G,
\end{equation}
evaluated at \(\widehat\theta_\psi\). At a converged inner solution, the hypergradient is evaluated by solving a linear system involving the inner Hessian. It is exact up to the tolerance of the inner optimization and assumes that the fitted solution is locally isolated with a positive-definite Hessian. Since the negative-binomial objective need not be globally convex, we use multiple starting points and report the smallest Hessian eigenvalue at each final solution as a diagnostic.

The continuous components of \(\psi\) are optimized jointly within prespecified bounds. The number of annual harmonics is treated as a discrete choice, with the search repeated for each candidate value. Trend, weekly, covariate, and GP-correction blocks are removed in practice when their scales reach the lower end of the search range. The selected model is then refitted on the full fitting period before the component projection and held-out evaluation. The validation segment is the final 40\% of the fitting period, and both the split and random seed are fixed. Appendix~C gives the starting-point design, optimization budgets, derivative checks, and runtimes.

The same validation window and one-step predictive criterion are used for the differentiable count-process benchmarks. Neural models are selected on the same window using their stated grid, while the random-histogram model integrates over its posterior rather than selecting a single hyperparameter value. The common criterion improves comparability, although the parameterizations and search procedures remain model-specific. We use forward validation rather than a Laplace marginal-likelihood criterion because it applies to all of the fitted model classes and directly targets the quantity used for evaluation.

\paragraph{Fitting under the stability constraint.}
For GP-DHP we impose \(R_+\leq 1-\delta\), with \(\delta=10^{-4}\), during hyperparameter selection. The inner constrained MAP problem is solved by an augmented-Lagrangian method, while the outer objective remains the forward-validation score in Equation~\eqref{eq:fv-criterion}. The implicit hypergradient is evaluated at the constrained inner solution. Appendix~C gives the multiplier updates, derivative calculation, and numerical checks. We report the achieved value of \(R_+\) for each fit; all reported fits satisfy \(R_+<1\).

\subsection{Decomposition into Baseline and Excitation Components}\label{sec:decomposition}

While inference is performed on the latent trajectory \(\boldsymbol{\ell}^*\), recovering the individual components \(b(t)\) and \(f(d)\) is desirable for interpretability, particularly in epidemiological or social applications, where separating background variation from residual lag dependence provides interpretable summaries rather than causally identified mechanisms. We recover the minimum-prior-norm decomposition of the centered latent trajectory into a baseline vector \(b\) and a fitted GP excitation contribution \(u\)---the counterpart of the prior GP correction \(u(d)\) of Equation~\eqref{eq:bridge-prior}, sharing its covariance \(K_g\)---and then add back the negative-binomial kernel.

Let
\[
\boldsymbol{z}^*=\boldsymbol{\ell}^*-Xm_{\mathrm{NB}},
\qquad
\qquad
K_z=K_b+XK_gX^\top.
\]
The decomposition is the solution of
\begin{equation}\label{eq:bridge-decomp-optimization}
\min_{b,u}
\left\{
\frac{1}{2}b^\top K_b^{-1}b+\frac{1}{2}u^\top K_g^{-1}u
\right\}
\quad\text{subject to}\quad
\boldsymbol{z}^*=b+Xu.
\end{equation}
The recovered excitation is then \(\widehat f=m_{\mathrm{NB}}+\widehat u\).

\begin{Proposition}[Component projection]
\label{prop:hard-constraint}
Let \(K_b\in\mathbb{R}^{T\times T}\) and \(K_g\in\mathbb{R}^{D_{\max}\times D_{\max}}\) be symmetric positive definite, let \(X\in\mathbb{R}^{T\times D_{\max}}\), and let \(\boldsymbol{z}^*\in\mathbb{R}^T\). Consider
\[
\min_{b,u}
\frac{1}{2}b^\top K_b^{-1}b+\frac{1}{2}u^\top K_g^{-1}u
\quad\text{subject to}\quad
\boldsymbol{z}^*=b+Xu.
\]
Define \(K_z=K_b+XK_gX^\top\in\mathbb{R}^{T\times T}\). Then:
\begin{enumerate}
\item The feasible set is nonempty and the objective is strictly convex on it. Hence there is a unique minimizer.
\item The unique minimizer \((\widehat b,\widehat u)\) is
\[
\widehat b=K_bK_z^{-1}\boldsymbol{z}^*,
\qquad
\widehat u=K_gX^\top K_z^{-1}\boldsymbol{z}^*.
\]
\item The minimum value equals \(\tfrac{1}{2}\boldsymbol{z}^{*\top}K_z^{-1}\boldsymbol{z}^*\).
\end{enumerate}
\end{Proposition}

\textbf{Interpretation.}
Proposition~\ref{prop:hard-constraint} is a standard constrained Gaussian projection (the unique minimum-norm split, in the norms induced by \(K_b\) and \(K_g\), that exactly reconstructs the centered trajectory), and its short KKT proof is deferred to Appendix~A. We state it not as an optimization result but for its statistical reading, which is what makes the collapsed fit interpretable. If \(b\sim\mathcal{N}(0,K_b)\) and \(u\sim\mathcal{N}(0,K_g)\) are the independent component priors and \(z=b+Xu\) their collapsed sum, then \(\operatorname{Cov}(b,z)=K_b\), \(\operatorname{Cov}(u,z)=K_gX^\top\), and \(\operatorname{Var}(z)=K_z\), so the formulas of the proposition are exactly
\[
\widehat b=\mathbb{E}\bigl(b\mid z=\boldsymbol{z}^*\bigr),
\qquad
\widehat u=\mathbb{E}\bigl(u\mid z=\boldsymbol{z}^*\bigr):
\]
the projection is the conditional mean of each component given the fitted collapsed trajectory, and the minimum value \(\tfrac12\boldsymbol{z}^{*\top}K_z^{-1}\boldsymbol{z}^*\) is the collapsed prior energy. Once the collapsed representation of Section~\ref{Inference} exists, the decomposition therefore follows from it with no further modelling choices. The final excitation summary is
\[
\widehat f=m_{\mathrm{NB}}+\widehat u.
\]

In the fitted model the component covariances are finite-rank (\(K_b=BB^\top\), \(K_g=LL^\top\)), and the projection reduces to the direct coefficient read-off \(\widehat b=B\theta_b^*\), \(\widehat u=L\theta_g^*\) from the fitted MAP coefficient \(\theta^*\); Appendix~A shows this coincides with Proposition~\ref{prop:hard-constraint} in the degenerate limit and holds for every reported fit.

\subsection{Computational Complexity and Efficiency}
\label{sec:computation}

A key advantage of the collapsed representation is that it exposes the covariance structure of the latent trajectory. Classical evaluations of the excitation term in Equation~\eqref{eq:discrete-hawkes} can require nested history summations when no truncation or convolution structure is used. In the collapsed representation, temporal dependence is encoded in the GP prior covariance, while the observation log-likelihood is a sum over time points depending only on \(\ell(t)\). Thus MAP inference reduces to optimization of a latent GP posterior with covariance
\[
K_z=K_b+XK_gX^\top.
\]

The saving comes from this structure. The excitation contribution \(XK_gX^\top\) reduces to causal convolutions with the count sequence, which we evaluate together with their adjoints by fast-Fourier-transform convolution, and both the baseline and lag covariances admit fast structured matrix--vector products. A single objective, gradient, or hypergradient evaluation therefore costs \(O(T\log T+D_{\max}^2)\), near-linear in the series length \(T\), with no \(T\times T\) factorization at any point. The structured products for the baseline and lag covariances, and the large-\(D_{\max}\) structured-kernel option, are collected in Appendix~C.

\paragraph{Projection summaries.}
The decomposition in Proposition~\ref{prop:hard-constraint} is deterministic once the latent MAP trajectory and kernel hyperparameters have been selected. In the simulation studies below, uncertainty bands are therefore reported as between-replicate summaries rather than local posterior credible intervals. For real-data applications, the projected baseline and excitation should be interpreted as regularized component summaries of the fitted latent intensity, while predictive performance is assessed by one-step-ahead held-out log-likelihood.

\paragraph{Identifiability and regularization.}
Because \(b(t)\) and \(\sum_dN(t-d)f(d)\) enter additively, the decomposition is necessarily prior-dependent. The separated level, trend, and finite-Fourier components in \(K_b\) make persistent and calendar-scale variation natural to the baseline, while the excitation prior makes a standard short-lag Hawkes memory kernel available without excluding GP departures from that kernel. The projection above then selects the exact decomposition with minimum combined prior norm. This is the sense in which the model regularizes the otherwise ambiguous separation of baseline and excitation.

\section{Experiments}\label{sec:experiments}

\subsection{Synthetic Data}\label{synthetic}

We evaluate the ability of GP-DHP to recover latent self-exciting dynamics by simulating from known discrete Hawkes models and assessing whether the model can reconstruct both the baseline and excitation components from observed count data. Neither the baseline nor excitation functions are assumed known during inference.

\paragraph{Model Specification}

Each simulated time series is generated from a discrete Hawkes process of length \( T = 6000 \). The event intensity is composed of a baseline function \( b(t) \) and an excitation kernel \( f(d) \), as described in Section~\ref{Proposed Model}. We fit the GP-DHP model to each time series, where the baseline prior uses the separated level, linear, and finite-Fourier covariance from Equation~\eqref{eq:Kb-fourier},
\[
K_b = \sigma_{\mathrm{level}}^2 \mathbf 1\mathbf 1^\top
      + \sigma_{\mathrm{lin}}^2 \mathbf{t}\mathbf{t}^\top
      + \sigma_{\mathrm{season}}^2 H_KH_K^\top,
\]
where \(H_K\) contains sine and cosine seasonal harmonics but no intercept column; this is the probabilistic covariance \(K_b\), the numerical jitter entering only the factorized \(K_b^{\mathrm{num}}=K_b+\epsilon_b^2 I\), not the fitted model.

The excitation kernel has the form in Equation~\eqref{eq:bridge-prior}: a normalized negative-binomial kernel plus a GP correction. The correction has the nonstationary lag covariance
\[
a(d) \;=\; \sigma_g \exp\!\Big(-\tfrac{\beta d}{2}\Big),
\qquad
w(d) \;=\; \frac{1 - e^{-\beta d}}{\beta\,\ell_g},
\]
\[
K_g(d, d') \;=\; a(d)\,a(d') \;
\exp\!\left( -\tfrac{1}{2}\,\big(w(d)-w(d')\big)^2 \right) + \epsilon_g^2\delta_{dd'}.
\]
This construction smoothly correlates nearby lags while attenuating long-range effects through \(a(d)\) and compressing large lags via \(w(d)\). Inference proceeds over the latent additive intensity, followed by the closed-form projection to $b$ and $f$ (see Section~\ref{Inference}). The diagonal terms in \(K_b\) and \(K_g\) are fixed numerical stabilizers and are not selected as modelling hyperparameters.

\paragraph{Selection for the synthetic study.}

For each synthetic dataset, GP-DHP hyperparameters are selected by the forward-validation selection of Section~\ref{sec:hyperparameter-selection}. After selection, the projected MAP components are displayed. In replicated experiments, shaded bands denote empirical pointwise ranges across independent simulations, not posterior credible intervals. The search is confined to the ranges in Supplementary Table~\ref*{tab:hyper-ranges}, which act as box constraints for the multi-start optimizer; the optimizer explores this domain by forward-validated predictive log-likelihood and does not enumerate the product of the listed values (which would be on the order of \(10^8\) combinations).

\paragraph{Synthetic baseline regimes}
The synthetic experiments below use three nonstationary baseline regimes:
\[
\begin{aligned}
b_{\mathrm{lin}}(t) &= 0.5 + 0.001t,\\
b_{\mathrm{sea}}(t) &= 0.8 + 0.05\cos(2\pi t/365) + 0.25\sin(2\pi t/365),\\
b_{\mathrm{lin+sea}}(t) &= 0.5 + 0.001t + 0.05\cos(2\pi t/365) + 0.25\sin(2\pi t/365).
\end{aligned}
\]
These correspond to a linearly increasing baseline, a seasonal baseline, and a combined linear--seasonal baseline. The two synthetic studies below use these regimes in complementary ways: Figure~\ref{fig:excitation-comparison} focuses on excitation-kernel recovery, while Figure~\ref{fig:baseline_excitation_comparison} focuses on baseline recovery.

\subsubsection{Simulation Design - Excitation Function Recovery}

\begin{figure}[t]
    \centering
    \includegraphics[width=0.8\textwidth]{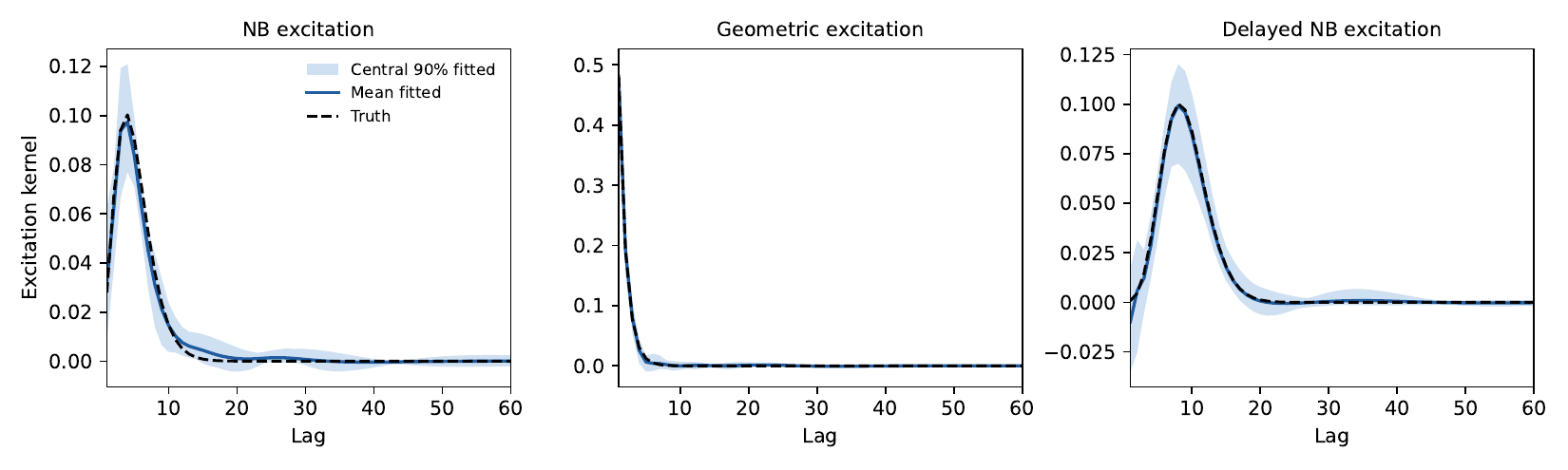}
    \caption{Excitation-kernel recovery in three representative synthetic DHP simulations. The panels compare the true excitation kernel with the projected MAP estimate from GP-DHP for negative-binomial, geometric, and delayed negative-binomial excitation settings. Black curves denote the true excitation kernels; grey curves denote mean projected MAP estimates across 10 simulations; shaded bands give central \(90\%\) pointwise between-replicate ranges.}
    \label{fig:excitation-comparison}
\end{figure}

We first assess whether the excitation kernel can be recovered in representative settings where both the baseline dynamics and the excitation shape vary. The three scenarios shown in Figure~\ref{fig:excitation-comparison} use the baseline regimes above together with three qualitatively different excitation kernels: a negative-binomial kernel, a geometric kernel, and a delayed negative-binomial kernel. The excitation settings are
\[
\begin{array}{ll}
\text{Negative binomial:} & K_{\mathrm{NB}}=0.6,\ (r,p_{\mathrm{lag}})=(6,0.6),\\
\text{Geometric:} & K_{\mathrm{NB}}=0.8,\ (r,p_{\mathrm{lag}})=(1,0.6),\\
\text{Delayed negative binomial:} & K_{\mathrm{NB}}=0.8,\ \text{mean lag }8,\ \text{size }25,
\end{array}
\]
with \(q_{\mathrm{NB}}(\cdot;r,p_{\mathrm{lag}})\) the kernel of Equation~\eqref{eq:m-nb}, supported on lags \(d\ge1\); the geometric kernel is its \(r=1\) special case. The delayed negative-binomial setting is included to check recovery of a non-monotone excitation kernel whose peak occurs after several time steps, rather than at the first lag. For each setting, we simulate 10 independent datasets of length \(T=6000\), select hyperparameters by the forward-validation criterion of Section~\ref{sec:hyperparameter-selection}, and compare the projected MAP estimate of the excitation kernel \(\hat f(d)\) with the ground truth.

Figure~\ref{fig:excitation-comparison} shows that the fitted kernels recover the main excitation shapes in these three cases. The negative-binomial and geometric kernels are recovered closely, and the delayed negative-binomial setting recovers the location and shape of the delayed peak.

\subsubsection{Simulation Design - Baseline Function Recovery}

To assess recovery of nonstationary baseline structure, we conduct a complementary experiment in which the baseline varies across the three regimes defined above, while the excitation kernel is held fixed in the data-generating process (but still estimated). Specifically, we simulate datasets of length \(T=6000\), each using the same normalized negative-binomial excitation kernel
\[
f(d)=0.5\,q_{\mathrm{NB}}(d;r=6,p_{\mathrm{lag}}=0.6), \qquad d=1,\ldots,D_{\max},
\]
with NB2 observation size \(\kappa=100\). We simulate 10 independent datasets under each baseline setting.

\begin{figure}[t]
    \centering
    \includegraphics[width=0.8\textwidth]{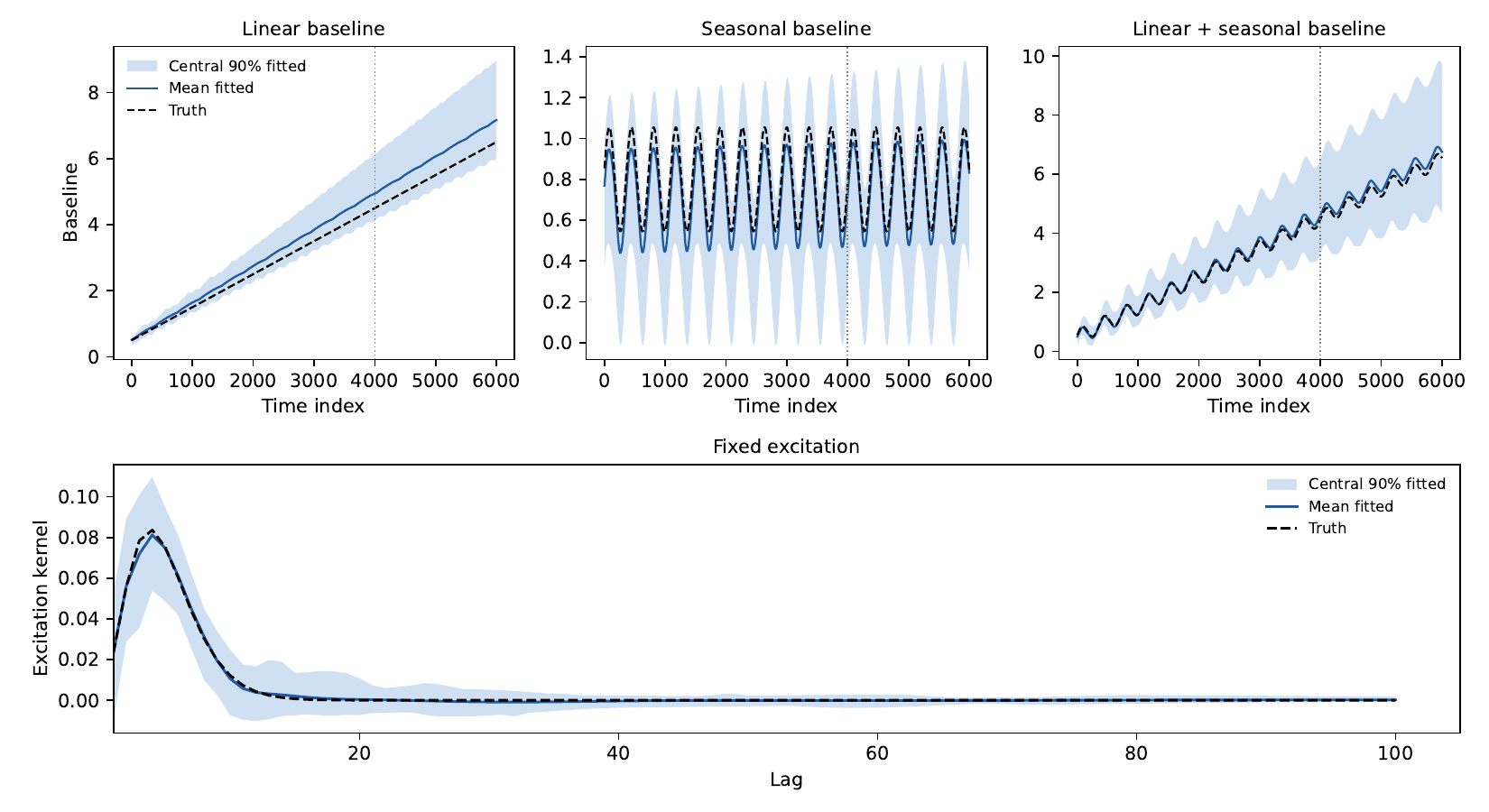}
    \caption{Recovery of baseline and excitation components under three nonstationary baseline settings. The excitation kernel is held fixed and the baseline varies across a linear trend, a seasonal baseline, and a combined linear--seasonal baseline. Black curves denote the true components; grey curves denote mean projected MAP estimates across 10 simulations; shaded bands give central \(90\%\) pointwise between-replicate ranges across simulations.}
    \label{fig:baseline_excitation_comparison}
\end{figure}

After fitting GP-DHP to each dataset, we apply the closed-form projection in Proposition~\ref{prop:hard-constraint} to recover the estimated baseline and excitation components. Figure~\ref{fig:baseline_excitation_comparison} presents the pointwise mean projected estimates, with central \(90\%\) pointwise between-replicate bands across the simulations. The recovered baselines track the main temporal structure in all three settings, while the estimated excitation functions remain stable across the different baseline regimes.

\subsubsection{Collapsed versus uncollapsed optimization}\label{sec:collapsed-vs-uncollapsed}

As a timing check on the collapsed representation, we simulated data from a linear--seasonal baseline
\[
b(t)=0.5+0.001t+0.05\cos(2\pi t/365)+0.25\sin(2\pi t/365)
\]
and an excitation kernel with a component beyond the fitted negative-binomial backbone,
\[
\begin{aligned}
f(d)={}&0.35\,q_{\mathrm{NB}}(d;\text{mean lag}=4,\text{size}=10)\\
&{}+0.20\,q_{\mathrm{NB}}(d;\text{mean lag}=12,\text{size}=25),\qquad d=1,\ldots,D_{\max}.
\end{aligned}
\]
We used \(D_{\max}=200\), three replicates at each \(T\in\{2000,5000\}\), an 80/20 train--test split, and NB2 observation size \(\kappa=100\). For each dataset we fitted GP-DHP twice at the same fixed hyperparameters: once using the collapsed optimization and once by direct optimization over the full latent \(b\) and \(u\). The direct uncollapsed fit used the same NB2 likelihood, rectified link, \(K_b\), \(K_g\), and lag matrix \(X\).

Table~\ref{tab:collapsed_uncollapsed} shows that the two fits give essentially the same fitted latent intensity and held-out predictive likelihood, while the collapsed optimization is about an order of magnitude faster in this fixed-hyperparameter check.

\begin{table}[t]
\centering
\small
\setlength{\tabcolsep}{4pt}
\begin{tabular}{rrcrrrrr}
\hline
\(T\) & \(D_{\max}\) & Replicates & Collapsed (s) & Direct (s) & Speedup & Max RMSE & Max \(|\Delta\mathrm{pLL}|/n_{\mathrm{test}}\) \\
\hline
2000 & 200 & 3/3 & 0.044 & 0.453 & \(10.37\times\) & \(4.13\times10^{-5}\) & \(9.29\times10^{-7}\) \\
5000 & 200 & 3/3 & 0.115 & 1.226 & \(11.62\times\) & \(1.36\times10^{-4}\) & \(8.81\times10^{-6}\) \\
\hline
\end{tabular}
\caption{Collapsed and direct full-latent optimization on synthetic DHP data at fixed hyperparameters. Times are medians over three replicates. The last two columns report the largest discrepancy across replicates in fitted latent intensity and held-out predictive log-likelihood per test observation.}
\label{tab:collapsed_uncollapsed}
\end{table}

\FloatBarrier
\subsection{Real Data}\label{sec:real-data}

We evaluate GP-DHP and benchmark discrete-time count models on five real-world count series: two weekly disease-surveillance series (Singapore dengue and German cryptosporidiosis), two daily shooting-count series (New York City and the Gun Violence Archive), and a daily worldwide terrorism-incident series (the RAND Database of Worldwide Terrorism Incidents). All fitted count models use a negative-binomial observation distribution for the final comparisons, and all GP-DHP hyperparameters are selected on the fitting period by the forward-validation criterion of Section~\ref{sec:hyperparameter-selection}. For the two daily shooting series we additionally supply GP-DHP and the count-process benchmarks with the external covariates described below. The final part of each series is held out as a test period and is used only for one-step-ahead predictive evaluation.

\paragraph{Datasets.}
The five data sets are chosen to cover both weekly disease-outbreak series and higher-count daily event streams, and to include series with and without useful external drivers:
\begin{itemize}
    \item \textbf{Singapore dengue.} Weekly laboratory-confirmed dengue case counts for Singapore, taken from the Ministry of Health Weekly Infectious Disease Bulletin as distributed through \texttt{data.\allowbreak gov.\allowbreak sg} \citep{moh_dengue_sg}. Dengue is a mosquito-borne viral infection that is endemic in Singapore, with pronounced multi-year epidemic cycles and a within-year peak in the warmer mid-year months. The series is high-count, strongly overdispersed, and strongly self-exciting, with sharp outbreak waves that build up and decay over several weeks. We use consecutive epidemiological weeks from 2012 onward.
    \item \textbf{German cryptosporidiosis.} Weekly notified cryptosporidiosis case counts for Germany, obtained from the Robert Koch Institute surveillance portal SurvStat@RKI~2.0 \citep{survstat_rki}. This is the same national reporting source as the shorter series used in earlier work; here we extract a longer, up-to-date national series. Cryptosporidiosis is a gastrointestinal disease caused by the protozoan parasite \textit{Cryptosporidium}, transmitted substantially through contact with farm animals and the rural environment \citep{hunter2003epidemiology}, with low endemic levels interrupted by seasonal late-summer outbreaks \citep{yoder2010cryptosporidiosis}. We use the pre-pandemic weekly series (through 2019) to avoid the COVID-era reporting disruption.
    \item \textbf{NYC shootings.} The NYC shootings series contains daily shooting-incident counts for New York City, taken from the NYPD shooting-incident data published through NYC Open Data \citep{nycopendata_shootings_2006_present}. It provides a higher-count urban event series with strong calendar structure and short-term dependence. Because daily shootings are known to co-vary with weather and the calendar, this series is paired with external covariates (below).
    \item \textbf{GVA shootings.} The GVA series aggregates daily shooting counts from the Gun Violence Archive, using the 2014--2023 all-shootings dataset \citep{uma2024_gva_all_shootings}, available from figshare at \url{https://doi.org/10.6084/m9.figshare.25517224.v2}; the Gun Violence Archive collection methodology is documented at \url{https://web.archive.org/web/20260331020805/https://www.gunviolencearchive.org/methodology}. It is a larger-scale daily event stream with substantially higher counts than the disease series, and is likewise paired with external covariates.
    \item \textbf{Worldwide terrorism (RAND).} Daily counts of worldwide terrorist incidents from the RAND Database of Worldwide Terrorism Incidents \citep{rand_terrorism}, formed by aggregating all recorded incidents to their occurrence date. This is a long, high-volume daily event stream ($40{,}129$ incidents over 1968--2009) with strong self-excitation from sustained campaigns and reprisals and pronounced day-of-week structure. It is markedly non-stationary: worldwide activity is comparatively sparse through the 1970s--1990s and rises sharply during the mid-2000s, so that the held-out final five years form a substantially higher-intensity regime than the fitting period, making this a deliberately demanding out-of-regime forecasting test. The series uses no external covariates.
\end{itemize}

\paragraph{External covariates for the daily series.} For the two daily shooting series we form the covariate block of Equation~\eqref{eq:cov-baseline} from a daily \emph{temperature anomaly} (the local daily mean temperature minus a smooth harmonic climatology, standardized on the fitting period, which isolates unusually warm or cool days from the seasonal cycle) and a set of per-holiday binary indicators for nine major U.S. public holidays, each with its own coefficient rather than a single pooled effect. The temperature anomaly carries one prior scale and the holiday block a shared scale, both selected by forward validation (Section~\ref{sec:hyperparameter-selection}); the weekly disease series use no external covariates.

Table~\ref{tab:realdata-datasets} summarizes the fitting and test periods. Each model is fit once on the fitting period (used for both fitting and hyperparameter selection) and is not refit within the held-out test window; sequential one-step scoring lets earlier test observations enter the history for later predictions but excludes the current observation from its own forecast. For GP-DHP the one-step predictive mean at a held-out time applies the softplus link to the fitted Fourier baseline, the covariate term (daily series only), and the lagged-count excitation evaluated on observations prior to \(t\), exactly as in the model intensity of Section~\ref{sec:model}.

\FloatBarrier
\begin{table}[t]
\centering
\small
\resizebox{\textwidth}{!}{%
\begin{tabular}{lllllrr}
\hline
\textbf{Dataset} & \textbf{Frequency} & \textbf{Fitting period} & \textbf{Test period} & \textbf{Cov.} & \textbf{Season} & \textbf{Max. lag} \\
\hline
Singapore dengue & Weekly & 314 weeks (2012--2017) & 260 weeks (2018--2022) & -- & 52 & 100 \\
German cryptosporidiosis & Weekly & 730 weeks (2001--2014) & 260 weeks (2015--2019) & -- & 52 & 100 \\
NYC shootings & Daily & 6209 days (2006--2022) & 1096 days (2023--2025) & temp\,+\,hol. & 365 & 100 \\
GVA shootings & Daily & 2922 days (2014--2021) & 730 days (2022--2023) & temp\,+\,hol. & 365 & 100 \\
Worldwide terrorism (RAND) & Daily & 13476 days (1968--2004) & 1826 days (2005--2009) & -- & 365 & 100 \\
\hline
\end{tabular}%
}
\caption{Real-data series and evaluation splits. The fitting period is used for model fitting and hyperparameter selection; the test period is held out for sequential one-step-ahead predictive scoring. ``Cov.''\ indicates the external covariate block (deseasonalized temperature anomaly and per-holiday indicators) supplied to the daily shooting series; the weekly disease series use no external covariates. ``Season''\ is the annual seasonal period; the daily models additionally include a weekly ($P_w=7$) baseline component.}
\label{tab:realdata-datasets}
\end{table}

We compare GP-DHP against nine discrete-time count-process benchmarks, all using the same negative-binomial observation model and, where applicable, the same covariate design and forward-validation split. The benchmarks comprise a baseline-only model; four parametric DHPs with constant, linear, sinusoidal, or linear-plus-sinusoidal baselines; NB-INGARCH; and three flexible-excitation models based on fixed-bin, random-histogram, and GP-modulated kernels. The random-histogram model follows the construction of \citet{browning2022}, adapted to the common observation and scoring framework, while the discrete GP-Hawkes is motivated by the GP-modulated kernels of \citet{zhou2020} and \citet{zhang2020}. Full specifications are given in the Supplementary Material, and selected settings in Supplementary Table~\ref*{tab:supp-extra-benchmark-settings}.

Neural autoregressive count models fitted under the same evaluation protocol are reported in Supplementary Table~\ref*{tab:supp-neural}. They were less accurate on these series and do not provide the baseline--excitation decomposition considered here. Continuous-time neural point-process models such as Neural Hawkes \citep{mei2017} require exact event times and are therefore not directly applicable to the binned-count data studied here.

Each model is scored by one-step-ahead predictive log-likelihood (pLL) on the held-out test period---the summed negative-binomial one-step log-scores \(\log p(N(t)\mid\mathcal{H}(t-1))\), including all normalizing constants---as a plug-in score conditional on the fitted coefficients and selected hyperparameters. Because total-pLL gaps can be small relative to the number of test points, we compare each benchmark's per-observation log-score sequence against GP-DHP with a Diebold--Mariano test \citep{diebold1995comparing} and control the family-wise error rate with a Holm correction within each fitted cell (a dataset--covariate combination); the exact statistic and the full protocol are given in the Supplementary Material. We report a benchmark as not significantly different when its adjusted \(p\)-value exceeds \(0.05\), rather than counting a small raw margin as a win; a nonsignificant test indicates only a failure to reject equal expected loss, not established equivalence. Additional held-out error metrics for GP-DHP, selected model settings, and held-out predictive-interval plots are reported in the supplementary material.

\FloatBarrier
\begin{table}[t]
\centering
\small
\begin{tabular}{lrrr}
\hline
\textbf{Model} & \textbf{Dengue} & \textbf{Crypto.} & \textbf{Terrorism} \\
\hline
\textbf{GP-DHP}         & \textbf{-1315.9} & \textbf{-947.3} & \textbf{-5079.5} \\
\hline
Baseline only           & $-1929.9^{***}$ & $-1009.3^{***}$ & $-7257.6^{***}$ \\
Discrete DHP            & $-1347.5^{**}$  & $-967.3^{***}$  & $-5125.0^{***}$ \\
Linear DHP              & $-1388.0^{***}$ & $-963.6^{**}$   & $-5387.3^{***}$ \\
Sinusoidal DHP          & $-1347.0^{**}$  & $-968.3^{***}$  & $-5127.9^{***}$ \\
Linear + Sinusoidal DHP & $-1393.9^{***}$ & $-956.9^{*}$    & $-5553.2^{***}$ \\
Histogram DHP-NB        & $-1338.6^{*}$   & $-957.1^{*}$    & $-5134.7^{***}$ \\
Random-histogram DHP-NB & $-1333.2^{*}$   & $-961.2^{**}$    & $-5230.1^{***}$ \\
GP-Hawkes (discrete)    & $-1357.1^{***}$ & $-947.0$        & $-5132.2^{***}$ \\
NB-INGARCH              & $-1380.1^{***}$ & $-963.0^{**}$   & $-5163.9^{***}$ \\
\hline
\end{tabular}
\caption{Held-out one-step-ahead negative-binomial predictive log-likelihoods on the three series without external covariates (higher is better). GP-DHP is shown in bold. Superscripts give Holm-adjusted Diebold--Mariano two-sided \(p\)-values for the per-observation gap to GP-DHP: \(^{*}p<0.05\), \(^{**}p<0.01\), \(^{***}p<0.001\); a benchmark with no superscript is not significantly different from GP-DHP at the 5\% level. All models use the same data partition and one-step predictive scoring rule.}
\label{tab:realdata-predictive-nocov}
\end{table}

\begin{table}[t]
\centering
\small
\setlength{\tabcolsep}{5.5pt}
\begin{tabular}{lrrrr}
\hline
\textbf{Model} & \multicolumn{2}{c}{\textbf{NYC shootings}} & \multicolumn{2}{c}{\textbf{GVA shootings}} \\
               & \textbf{no cov.} & \textbf{+\,cov.} & \textbf{no cov.} & \textbf{+\,cov.} \\
\hline
\textbf{GP-DHP}         & \textbf{-1965.7} & \textbf{-1949.3} & \textbf{-3055.9} & \textbf{-2985.0} \\
\hline
Baseline only           & $-2076.5^{***}$ & $-2059.8^{***}$ & $-3638.8^{***}$ & $-3583.4^{***}$ \\
Discrete DHP            & $-2010.5^{***}$ & $-2001.7^{***}$ & $-3285.4^{***}$ & $-3264.0^{***}$ \\
Linear DHP              & $-2007.9^{***}$ & $-1994.8^{***}$ & $-3304.3^{***}$ & $-3265.6^{***}$ \\
Sinusoidal DHP          & $-2008.4^{***}$ & $-2009.4^{***}$ & $-3286.2^{***}$ & $-3293.1^{***}$ \\
Linear + Sinusoidal DHP & $-2006.9^{***}$ & $-1997.0^{***}$ & $-3329.6^{***}$ & $-3258.8^{***}$ \\
Histogram DHP-NB        & $-1985.9^{***}$ & $-1973.4^{***}$ & $-3200.8^{***}$ & $-3122.5^{***}$ \\
Random-histogram DHP-NB & $-1992.3^{***}$ & $-1979.8^{***}$ & $-3104.5^{***}$ & $-3030.4^{***}$ \\
GP-Hawkes (discrete)    & $-1980.9^{***}$ & $-1968.9^{***}$ & $-3078.8^{**}$  & $-3014.6^{***}$ \\
NB-INGARCH              & $-1991.7^{***}$ & $-2028.6^{***}$ & $-3149.1^{***}$ & $-3101.2^{***}$ \\
\hline
\end{tabular}
\caption{Held-out one-step-ahead negative-binomial predictive log-likelihoods on the two daily shooting series, fitted without and with the external covariate block (deseasonalized temperature anomaly and per-holiday indicators). The bold and superscript conventions match Table~\ref{tab:realdata-predictive-nocov}.}
\label{tab:realdata-predictive-cov}
\end{table}

\paragraph{Predictive performance.}
Tables~\ref{tab:realdata-predictive-nocov} and~\ref{tab:realdata-predictive-cov} report held-out negative-binomial predictive log-likelihoods and pairwise Diebold--Mariano tests against GP-DHP. Wherever hyperparameter selection is required, models use the same validation window and one-step predictive criterion described in Section~\ref{sec:hyperparameter-selection}.

GP-DHP has the highest predictive log-likelihood on four of the five datasets. On German cryptosporidiosis the discrete GP-Hawkes is nominally ahead by less than half a nat, and the difference is not statistically significant after Holm adjustment (\(p=0.92\)). All other benchmark comparisons are significant at the 5\% level after adjustment. Thus, among the count-process models considered here, GP-DHP is either the best-performing model or is not significantly separated from the best on each series.

For the two shooting datasets, Table~\ref{tab:realdata-predictive-cov} also compares fits with and without external covariates. The covariates improve GP-DHP by 16 nats on NYC and 71 nats on GVA, and improve most of the other count-process models as well. GP-DHP remains the highest-scoring model in each case. Among the benchmarks, the flexible excitation models are generally the strongest competitors, although their relative performance varies across datasets: neither histogram specification dominates the other, and the discrete GP-Hawkes is the closest competitor on cryptosporidiosis. Held-out MAE and RMSE for GP-DHP are reported in Supplementary Table~\ref*{tab:supp-alt-metrics}; full pairwise score-test results are given with the predictive tables.

\paragraph{Calibration.}
Predictive log-likelihood measures the probability assigned to the observations but does not, by itself, assess calibration. Table~\ref{tab:realdata-calibration} therefore reports empirical coverage and mean width for 50\%, 80\%, and 95\% one-step-ahead predictive intervals from the fitted negative-binomial GP-DHP model.

The predictive intervals are generally conservative. Coverage is close to nominal for dengue and GVA, mildly conservative for cryptosporidiosis, and more clearly conservative for NYC shootings and worldwide terrorism, particularly at the lower nominal levels. Including the external covariates has little effect on coverage for either shooting series.

The aggregated probability-integral-transform diagnostic of \citet{czado2009} shows modest departures from uniformity for most series, with the clearest departure for cryptosporidiosis, where the predictive distribution tends to lie below the rising counts in the test period. Per-series PIT histograms and held-out predictive-interval plots are shown in Supplementary Figures~\ref*{fig:supp-pit} and~\ref*{fig:supp-predictive-intervals}.

\begin{table}[t]
\centering
\small
\setlength{\tabcolsep}{3.5pt}
\resizebox{\textwidth}{!}{%
\begin{tabular}{lrrrrrrrr}
\hline
\textbf{Dataset} & \textbf{Test $n$} & \textbf{$\kappa$} & \textbf{Cov. 50\%} & \textbf{Cov. 80\%} & \textbf{Cov. 95\%} & \textbf{Width 50\%} & \textbf{Width 80\%} & \textbf{Width 95\%} \\
\hline
Singapore dengue & 260 & 40.33 & 50.8 & 77.7 & 96.5 & 76.52 & 145.53 & 222.95 \\
German cryptosporidiosis & 260 & 16.78 & 56.5 & 83.8 & 94.6 & 12.93 & 24.65 & 37.85 \\
NYC shootings (with cov.) & 1096 & 16.33 & 70.2 & 93.0 & 99.5 & 2.19 & 4.25 & 6.24 \\
NYC shootings (no cov.) & 1096 & 13.40 & 70.3 & 92.2 & 98.8 & 2.17 & 4.22 & 6.19 \\
GVA shootings (with cov.) & 730 & 148.71 & 55.5 & 81.0 & 96.6 & 20.37 & 38.74 & 59.20 \\
GVA shootings (no cov.) & 730 & 104.18 & 56.7 & 82.5 & 96.8 & 22.30 & 42.36 & 64.81 \\
Worldwide terrorism (RAND) & 1826 & 6.88 & 62.8 & 88.3 & 97.9 & 6.79 & 12.94 & 19.90 \\
\hline
\end{tabular}%
}
\caption{Held-out predictive interval calibration for the proposed GP-DHP model. The two daily shooting series are shown both with and without covariates, matching the seven cells of the predictive tables; the weekly disease series and worldwide terrorism carry no covariate variant. Coverage entries are empirical percentages for central one-step-ahead predictive intervals; widths are mean interval widths on the count scale; \(\kappa\) is the selected negative-binomial size.}
\label{tab:realdata-calibration}
\end{table}

\subsection{Interpreting the fitted components}\label{sec:components}

The preceding results assess predictive accuracy and calibration. We now examine the fitted components of the final GP-DHP fits, since a main motivation for the model is to retain an interpretable decomposition of the fitted latent intensity into baseline and lagged-event contributions.

\begin{figure}
    \centering
    \includegraphics[width=\textwidth]{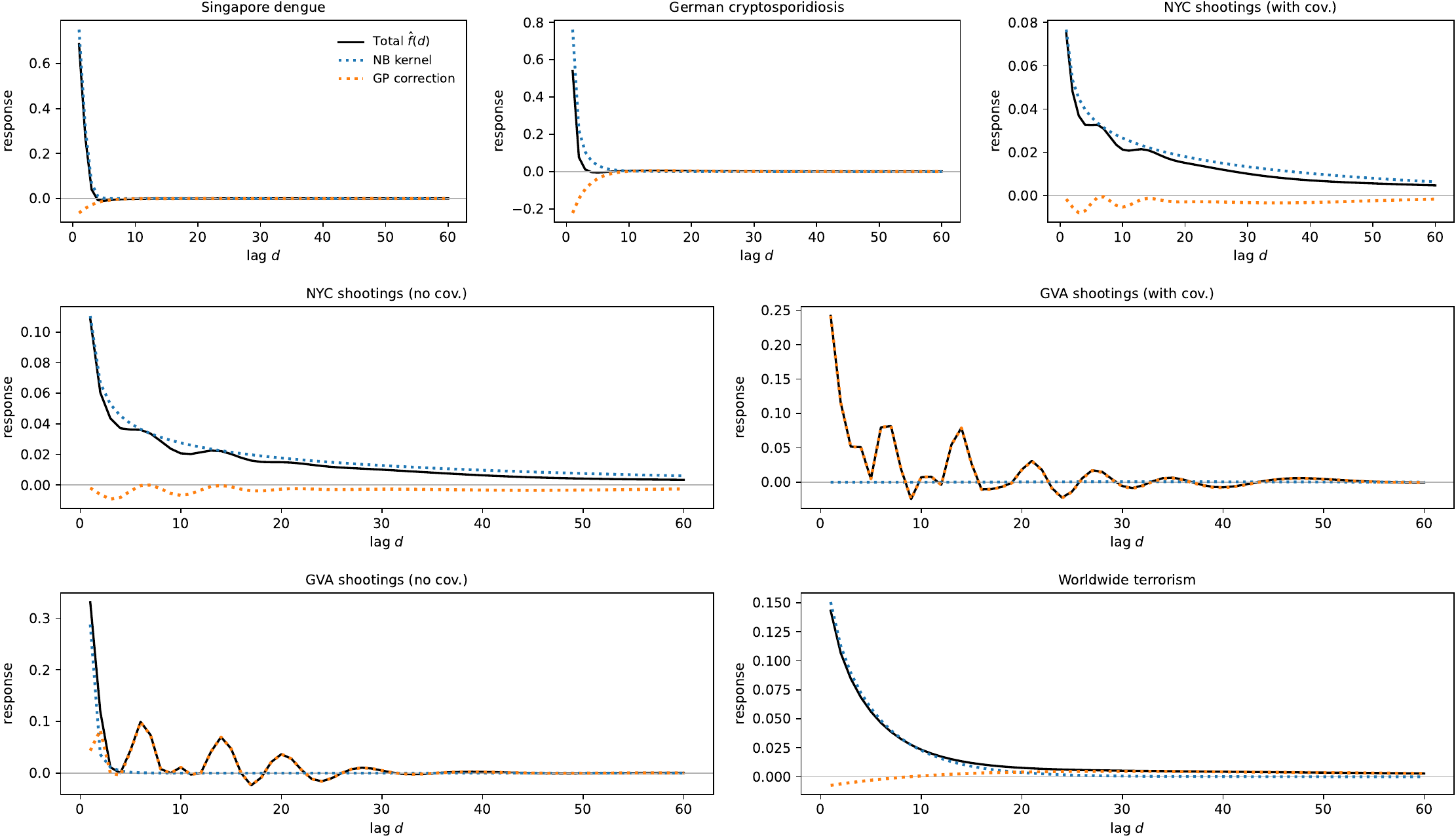}
    \caption{Fitted lag-response decompositions for the proposed GP-DHP on the seven fitted models---the five real-data series, with the NYC and GVA shooting series each shown with and without external covariates. Each panel shows the nonnegative negative-binomial kernel component, the GP correction, and their total fitted response over lags $1,\ldots,D_{\max}$. Negative GP corrections lower the latent response before the rectifying link and therefore do not imply negative conditional intensities.}
    \label{fig:realdata-kernel-small-multiples}
\end{figure}

Figure~\ref{fig:realdata-kernel-small-multiples} summarizes the fitted lag-response functions for the seven fitted GP-DHP models (the NYC and GVA shooting series are shown both with and without covariates; the two variants give visually similar kernels, as the covariates enter the baseline rather than the excitation). Both weekly disease series select sharply concentrated short-lag responses, peaking at lag one with most of their positive mass within two weeks (Supplementary Table~\ref*{tab:supp-excitation-summaries}), consistent with fast-building outbreak dynamics. The three daily series differ from one another: NYC shootings have a smooth, slowly decaying response with the longest memory, reaching out to about two months (half of its positive mass only by lag seventeen); GVA shootings have the shortest overall memory (half by lag six, ninety percent within three weeks) but carry a pronounced period-7 echo train, with bumps near lags 7, 14, and 21; and worldwide terrorism sits between them, rising as quickly as GVA (half its positive mass by lag six) but with a much longer tail that reaches to roughly seven weeks. The role of the GP correction varies across series: for Singapore dengue it drives the sharp lag-one spike; for German cryptosporidiosis it is a mild negative adjustment to an otherwise dominant negative-binomial kernel; for NYC shootings and worldwide terrorism it is essentially off, the smooth response coming from the negative-binomial backbone alone; and for GVA shootings it adds the short-lag structure on top of the backbone. The GP correction is added before the rectifying link, so negative values lower the latent lag response but do not imply negative fitted intensities.

The weekly (\(P_w=7\)) baseline block of Equation~\eqref{eq:Kb-fourier} spans the day-of-week means exactly, reducing the risk that deterministic weekly structure is assigned to the lagged-count term. Forward validation selects this block for all three daily series (\(\sigma_{\mathrm{week}}>0\); Supplementary Table~\ref*{tab:supp-selected-hyperparameters}), strongly for GVA and worldwide terrorism and weakly for NYC. After accounting for this weekly baseline, the fitted GVA excitation retains peaks near lags 7, 14, and 21, whereas the NYC and terrorism lag responses are smooth. Under the fitted regularized decomposition, this residual GVA pattern is assigned to lag dependence rather than deterministic day-of-week variation. Supplementary Table~\ref*{tab:supp-excitation-summaries} reports the positive-mass diagnostic, peak lag, and accumulation lags for each fitted excitation curve.

\begin{figure}
    \centering
    \includegraphics[width=\textwidth]{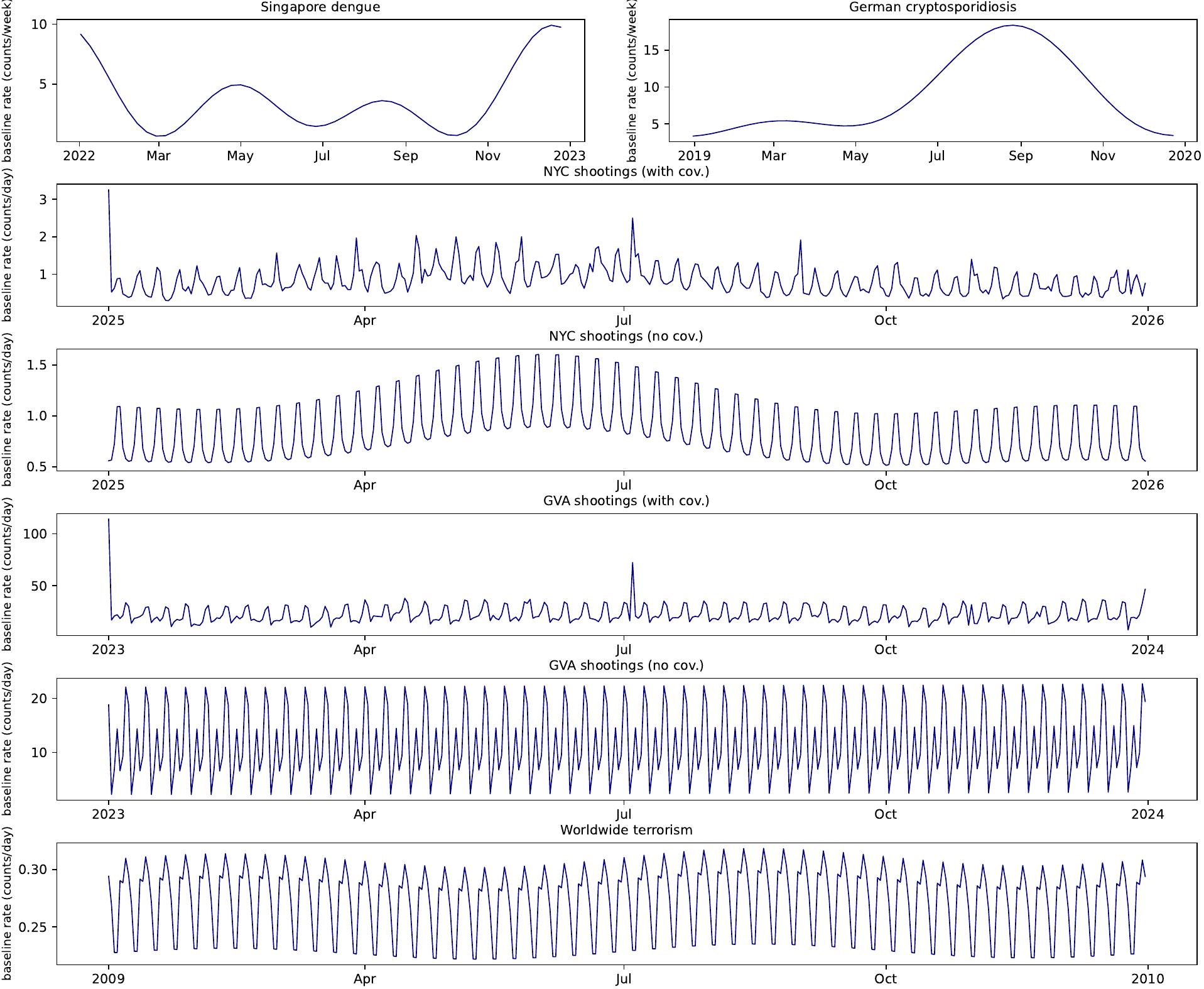}
    \caption{Fitted baseline rates for the proposed GP-DHP on the seven fitted models (the five real-data series, with the NYC and GVA shooting series each shown with and without external covariates). Each panel plots the excitation-free background intensity---the softplus of the projected baseline component \(b(t)\)---over the final year of the record (52 weeks for the weekly disease series, 365 days for the daily series), in the series' native per-period units (counts per week for the two disease series, per day for the shooting and terrorism series). Every test period is at least a year long, so the plotted window lies entirely within the held-out test span.}
    \label{fig:realdata-baseline-rates}
\end{figure}

The corresponding fitted baseline rates are shown in Figure~\ref{fig:realdata-baseline-rates}: for each of the seven fits it plots the excitation-free background intensity, the softplus of the projected baseline \(b(t)\), over the final year of each record in the series' native per-period units. This window covers one full seasonal period and lies entirely within the held-out test span, so the curves show the fitted background extrapolated to unseen data at a resolution where both the annual and the weekly structure are visible. The two weekly disease series show a pronounced annual cycle, peaking in late summer for cryptosporidiosis; the daily shooting series superimpose day-of-week modulation on the annual cycle, with a clear summer peak in the NYC envelope, and their covariate fits add sharp holiday excursions (visible on 1 January and 4 July) and day-to-day temperature-anomaly variation that the no-covariate fits cannot represent; the worldwide terrorism background reduces to a weekly cycle around a nearly constant level. Comparing the with- and without-covariate panels for NYC and GVA thus shows directly how much calendar and weather structure the covariates absorb into the baseline.

\begin{figure}
    \centering
    \includegraphics[width=0.95\textwidth]{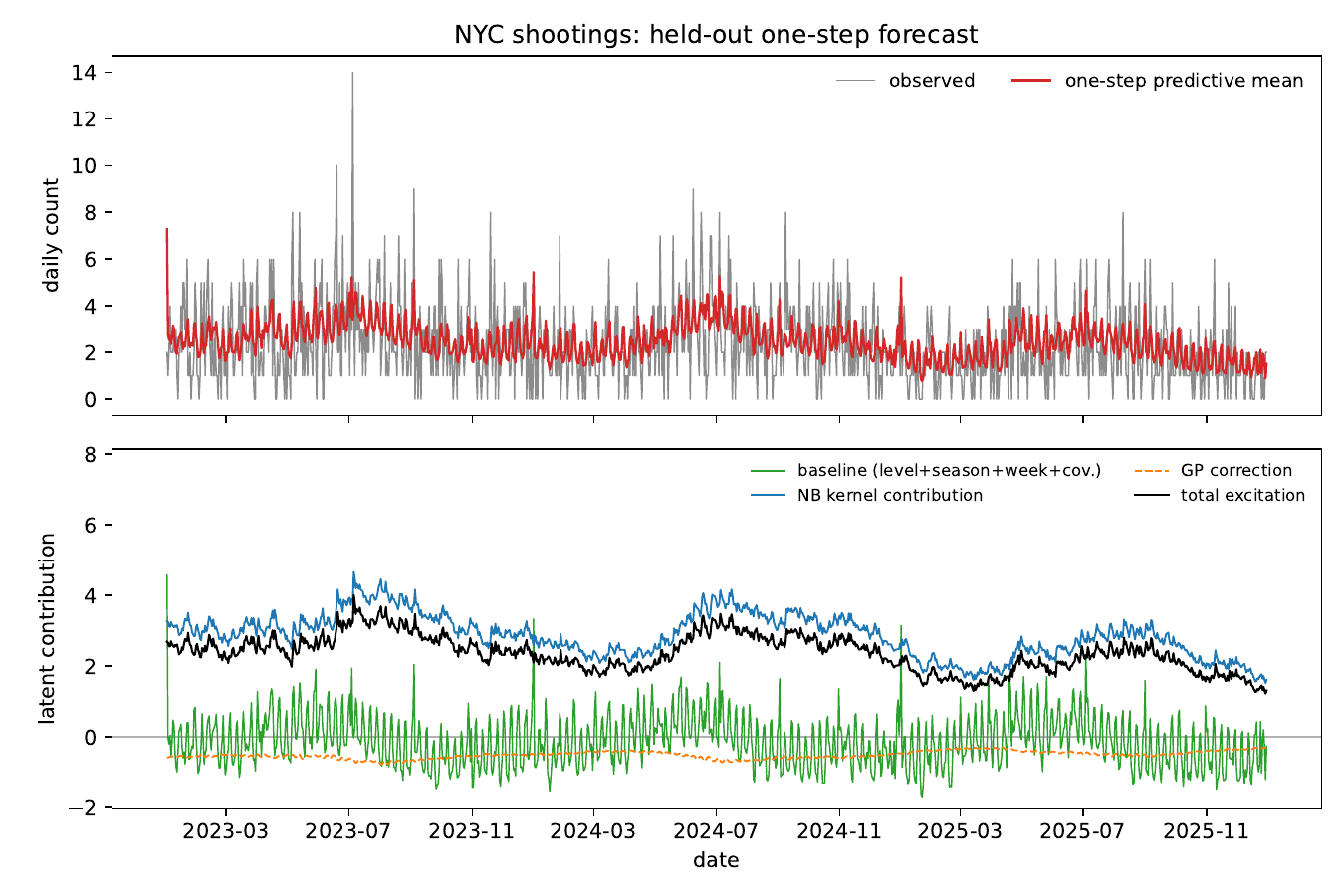}
    \caption{Projected components for the proposed GP-DHP fit \emph{with covariates} to held-out NYC shooting counts. The upper panel shows observed daily counts and the one-step predictive mean. The lower panel shows the additive components entering the latent intensity before the rectifying link: baseline (here including the day-of-week, holiday, and temperature-anomaly covariate contributions), negative-binomial kernel contribution, GP correction, and total excitation contribution.}
    \label{fig:nyc-components}
\end{figure}

Figure~\ref{fig:nyc-components} illustrates the fitted component decomposition for one representative daily series, NYC shootings (the with-covariate fit), on the held-out test period. The upper panel compares the observed daily counts with the one-step predictive mean. The lower panel separates the additive components entering the latent intensity before rectification: projected baseline, negative-binomial kernel contribution, GP correction, and total excitation contribution. Under the fitted regularized decomposition, the baseline captures the annual, day-of-week, holiday, and temperature-anomaly effects, while the lagged-count excitation carries much of the slower multi-week variation. This allocation depends on the model regularization and should not be interpreted causally.

\FloatBarrier

\section{Conclusion}\label{sec:conclusion}

We introduced GP-DHP, a semiparametric model for discrete-time Hawkes processes that combines flexible baseline and excitation components with a negative-binomial observation model. Marginalizing the GP components yields a collapsed latent process, and a finite-rank factorization permits MAP fitting and closed-form recovery of the fitted components. Across five applied datasets, GP-DHP attains the best predictive accuracy on four and is not significantly different from the best-performing model on the fifth. External covariates also improve the forecasts for the two daily shooting series.

The reported component curves are regularized MAP summaries and do not propagate posterior or sampling uncertainty to the fitted baseline and excitation. A Laplace approximation in \(\theta\) or a parametric bootstrap could provide this in future work. The model is most useful when the baseline is nonstationary or incompletely specified and the excitation kernel is unlikely to follow a fixed parametric form.

\section*{Reproducibility and Supplementary Materials}
An open-source implementation of GP-DHP is provided at \url{https://github.com/paulindani/GP-DHP}. It includes the collapsed-latent MAP fit, forward-validation hyperparameter selection, component projection, benchmark implementations, and scripts used to produce the tables and figures.

The repository also contains the five processed count series and the external covariates, with provenance notes recording their public sources: Singapore dengue from the Ministry of Health Weekly Infectious Disease Bulletin via \texttt{data.\allowbreak gov.\allowbreak sg} \citep{moh_dengue_sg}, German cryptosporidiosis from SurvStat@RKI~2.0 \citep{survstat_rki}, NYC shootings from NYC Open Data \citep{nycopendata_shootings_2006_present}, the Gun Violence Archive all-shootings dataset \citep{uma2024_gva_all_shootings}, and worldwide terrorism from the RAND Database of Worldwide Terrorism Incidents \citep{rand_terrorism}. The Supplementary Material reports held-out error metrics, selected hyperparameters and benchmark settings, excitation summaries, and held-out predictive-interval plots.


\bibliographystyle{plainnat}
\bibliography{bibliography}          

\clearpage
\setcounter{table}{0}
\setcounter{figure}{0}
\setcounter{equation}{0}
\renewcommand{\thetable}{S\arabic{table}}
\renewcommand{\thefigure}{S\arabic{figure}}
\renewcommand{\theequation}{S\arabic{equation}}

\begin{center}
{\Large Supplementary Material}
\end{center}

This supplementary material contains the proofs and implementation details supporting the main text (Appendices~A--C); full specifications of the benchmark models and of the evaluation protocol (Appendix~D); additional results for the five real-data experiments (Appendix~E); and the hyperparameter search ranges of the forward-validation selection, together with an illustration of the excitation prior (Appendix~F). All predictive values are computed from the predictions used in the main real-data comparison. Works cited in these appendices appear in the joint reference list above.

The source code of the experiments is available at \url{https://github.com/paulindani/GP-DHP/}.

\phantomsection
\section*{Appendix A: Proof of the component projection (Proposition~\ref*{prop:hard-constraint})}\label{app:proj-proof}

The proposition is a constrained Gaussian minimum-energy projection; we record the KKT calculation for completeness.

\begin{proof}[Proof of Proposition~\ref*{prop:hard-constraint}]
\textit{Feasibility and strict convexity.}
Feasibility holds since \((b,u)=(\boldsymbol{z}^*,0)\) satisfies \(\boldsymbol{z}^*=b+Xu\). The objective \(J(b,u)=\tfrac{1}{2}b^\top K_b^{-1}b+\tfrac{1}{2}u^\top K_g^{-1}u\) is strictly convex because \(K_b^{-1}\succ0\) and \(K_g^{-1}\succ0\). The feasible set is an affine subspace. Therefore a unique minimizer exists.

\textit{KKT conditions and solution.}
Form the Lagrangian, writing the multiplier as \(\nu\) to avoid a clash with the conditional intensity \(\lambda(t)\) (this \(T\)-dimensional multiplier is unrelated to the low-dimensional bilevel adjoint \(\boldsymbol\nu\) of Equation~(\ref*{eq:fv-hypergrad})),
\[
\mathcal{L}(b,u,\nu)=\tfrac{1}{2}b^\top K_b^{-1}b+\tfrac{1}{2}u^\top K_g^{-1}u+\nu^\top(\boldsymbol{z}^*-b-Xu),
\]
with multiplier \(\nu\in\mathbb{R}^T\). Stationarity gives
\[
\nabla_b\mathcal{L}=K_b^{-1}b-\nu=0 \Rightarrow b=K_b\nu,
\qquad
\nabla_u\mathcal{L}=K_g^{-1}u-X^\top\nu=0 \Rightarrow u=K_gX^\top\nu.
\]
Primal feasibility enforces
\[
\boldsymbol{z}^*-b-Xu
=
\boldsymbol{z}^*-K_b\nu-XK_gX^\top\nu=0
\Rightarrow K_z\nu=\boldsymbol{z}^*.
\]
For any nonzero \(v\in\mathbb{R}^T\),
\[
v^\top K_zv=v^\top K_bv+(X^\top v)^\top K_g(X^\top v)>0,
\]
since \(K_b\succ0\) and \(K_g\succ0\). Hence \(K_z\succ0\) and is invertible. Thus \(\nu=K_z^{-1}\boldsymbol{z}^*\), and substituting back gives the stated formulas. Uniqueness follows from strict convexity.

\textit{Minimum value.}
Using \(\widehat b=K_b\nu\) and \(\widehat u=K_gX^\top\nu\) with \(\nu=K_z^{-1}\boldsymbol{z}^*\),
\[
J(\widehat b,\widehat u)
=
\frac{1}{2}\nu^\top K_b\nu
+
\frac{1}{2}\nu^\top XK_gX^\top\nu
=
\frac{1}{2}\nu^\top K_z\nu
=
\frac{1}{2}\boldsymbol{z}^{*\top}K_z^{-1}\boldsymbol{z}^*.
\]
This completes the proof.
\end{proof}

\paragraph{Finite-basis (degenerate) specialization.} Proposition~\ref*{prop:hard-constraint} assumes positive-definite \(K_b,K_g\), whereas the fitted model uses the finite-rank component covariances \(K_b=BB^\top\) and \(K_g=LL^\top\), so \(K_z=AA^\top\) with \(A=[\,B\mid XL\,]\) is singular; we record how the MAP penalty and the projection specialize. Inference is carried out over the whitened coefficient \(\theta\) (Section~\ref*{sec:map-objective}): on the support \(\boldsymbol\ell\in Xm_{\mathrm{NB}}+\operatorname{col}(A)\) the collapsed prior penalty is the pseudo-inverse energy of the trajectory,
\[
\boldsymbol{z}^{*\top}K_z^{+}\boldsymbol{z}^*=\min_{\vartheta:A\vartheta=\boldsymbol{z}^*}\|\vartheta\|^2,\qquad K_z=AA^\top,
\]
with \(K_z^{+}\) the Moore--Penrose pseudo-inverse, so no \(T\times T\) system is formed. The reported decomposition is the coefficient-space split of the fitted \(\theta^*=(\theta_b^*,\theta_g^*)\), namely \(\widehat b=B\theta_b^*\) and \(\widehat u=L\theta_g^*\). When \(A\) has full column rank the representing coefficient is unique, so \(\theta^{*\top}\theta^*=\boldsymbol{z}^{*\top}K_z^{+}\boldsymbol{z}^*\) for \(\boldsymbol{z}^*=A\theta^*\); within the finite-basis parameterization \(b=B\theta_b\), \(u=L\theta_g\), the constraint \(\boldsymbol{z}^*=B\theta_b+XL\theta_g\) then has the unique solution \((\theta_b^*,\theta_g^*)\), so the component split is unique (as an unconstrained equation in \((b,u)\in\mathbb{R}^{T}\times\mathbb{R}^{D}\) it would not be), and it equals both the minimum-\(\|\theta\|^2\) decomposition above and the Gaussian conditional mean \(\mathbb{E}(b\mid z=\boldsymbol{z}^*)\) under the (possibly degenerate) Gaussian prior, defined by the Moore--Penrose conditioning formulas (\(K_z^{+}\) in place of \(K_z^{-1}\)). In this full-column-rank coefficient parameterization the degenerate conditioning formulas coincide directly with the coefficient-space split, so no nugget limit is invoked, and \(\epsilon_b\) does not enter the reported decomposition. For all reported fits \(A\) had full column rank, the smallest observed singular value being \(1.4\times10^{-2}\) (\hyperref[app:alfv]{Appendix~C}).

\setcounter{Lemma}{0}\renewcommand{\theLemma}{B.\arabic{Lemma}}
\phantomsection
\section*{Appendix B: Stability of the GP-DHP under \(R_+<1\)}\label{app:stability}

This appendix proves that the positive-mass condition \(R_+<1\) reported with every fit is sufficient for stability of the GP-DHP itself (negative-binomial observations, softplus link, deterministic baseline), not only for the Poisson--rectifier limit treated by \citet{costa2024general}, in two forms: geometric ergodicity under a constant baseline (Theorem~\ref*{thm:gpdhp-stability}), the idealized benchmark case; and, for an arbitrary deterministic baseline (seasonal, trended, or covariate-driven), mean growth of at most the order of the running maximum of the positive part of the baseline, hence uniformly bounded means whenever that path is bounded above (Proposition~\ref*{prop:growth}). Existence of the process at all times is automatic in discrete time and requires no condition on the kernel (Proposition~\ref*{prop:growth}); only the growth and mixing statements require \(R_+<1\). The proof follows the same Foster--Lyapunov route as \citet[Theorem~2.1]{costa2024general} in the Poisson--rectifier case: a linear Lyapunov function in the window coordinates with weights built from the positive part of the kernel, adapted here to the softplus link (Lemma~\ref{lem:link-envelope}), to a general observation law specified only through its conditional mean and the fact that it charges every point of \(\mathbb{N}_0\), and to a time-varying baseline (Proposition~\ref*{prop:growth}). We write \(\varrho\) for the Foster--Lyapunov \emph{drift coefficient} of Lemma~\ref{lem:weights}, a constant that controls the drift bound and depends on the arbitrary inflation \(\eta\), not an intrinsic mixing rate of the chain; it is distinct from the geometric convergence rate \(\bar\varrho\) of Theorem~\ref*{thm:gpdhp-stability} and unrelated to the augmented-Lagrangian penalty parameter \(\rho\) of \hyperref[app:alfv]{Appendix~C}. Throughout, the excitation kernel \(f\in\mathbb{R}^{D}\), \(D:=D_{\max}<\infty\), is fixed (as after any fit) and the model is that of Section~\ref*{sec:model}:
\begin{equation}\label{eq:app-model}
N(t)\mid\mathcal{H}(t-1)\;\sim\;\mathrm{NB}\{\lambda(t),\kappa\},\qquad
\lambda(t)=h_s\{\ell(t)\},\qquad
\ell(t)=b(t)+\sum_{d=1}^{D} f(d)\,N(t-d),
\end{equation}
with \(h_s(x)=s\log(1+e^{x/s})+\epsilon\), scale \(s>0\), floor \(\epsilon\ge0\) (the model of Section~\ref*{sec:model} fixes a small \(\epsilon>0\); every result below holds for any \(\epsilon\ge0\)), and dispersion \(\kappa\in(0,\infty)\); \(\mathrm{NB}\{\lambda,\kappa\}\) has mean \(\lambda\) and assigns positive probability to every point of \(\mathbb{N}_0\) whenever \(\lambda\in(0,\infty)\). Write \(x_+:=\max\{x,0\}\) and
\[
f^{+}_d:=\{f(d)\}_+,\qquad R_+:=\sum_{d=1}^{D}f^{+}_d .
\]
Only two features of the observation law are used: its conditional mean is \(\lambda(t)\), and it assigns positive probability to every point of \(\mathbb{N}_0\). In particular every statement below holds verbatim for Poisson observations (the \(\kappa\to\infty\) limit) and for every fixed \(\kappa\in(0,\infty)\); the overdispersion plays no role in the drift. The baseline \(b(t)\) is an arbitrary deterministic sequence; when the fitted model includes exogenous covariates, their contribution \(\sum_j c_j z_j(t)\) is folded into \(b(t)\), and every statement below is then conditional on the covariate path.

\begin{Lemma}[link envelope]\label{lem:link-envelope}
For every \(x\in\mathbb{R}\), \(s>0\) and \(\epsilon\ge0\),
\[
x_+ +\epsilon\;\le\;h_s(x)\;\le\;x_+ + s\log 2+\epsilon,
\]
and the upper bound is attained at \(x=0\).
\end{Lemma}

\begin{proof}
Since \(1+e^{x/s}\ge\max\{1,e^{x/s}\}\), monotonicity of the logarithm gives
\[
s\log(1+e^{x/s})\ge\max\{0,x\}=x_+,\qquad\text{hence}\qquad h_s(x)\ge x_++\epsilon\ge x_+.
\]
For \(x\le0\) we have \(1+e^{x/s}\le2\), so \(s\log(1+e^{x/s})\le s\log2=x_++s\log2\). For \(x>0\), the identity \(s\log(1+e^{x/s})=x+s\log(1+e^{-x/s})\) together with \(e^{-x/s}<1\) gives \(s\log(1+e^{x/s})\le x+s\log2\). Adding \(\epsilon\) yields the upper bound; at \(x=0\) both sides equal \(s\log2+\epsilon\).
\end{proof}

\begin{Lemma}[contraction weights]\label{lem:weights}
Let \(f^{+}\in[0,\infty)^{D}\) with \(R_+=\sum_{d=1}^D f^{+}_d<1\). Fix \(\eta\in\bigl(0,(1-R_+)/D\bigr)\) and set \(\tilde f^{\,+}_d:=f^{+}_d+\eta\). Then
(i) \(\varphi(\varrho):=\sum_{d=1}^{D}\tilde f^{\,+}_d\varrho^{-d}\) satisfies \(\varphi(\varrho)=1\) for exactly one \(\varrho\in(0,1)\); and
(ii) the weights
\[
\omega_j:=\varrho^{\,j-1}\sum_{i=j}^{D}\tilde f^{\,+}_i\varrho^{-i}\quad(j=1,\dots,D),\qquad \omega_{D+1}:=0,
\]
satisfy \(\omega_1=1\), \(\omega_j\ge\tilde f^{\,+}_j/\varrho\ge\eta/\varrho>0\), and
\begin{equation}\label{eq:app-weights}
\tilde f^{\,+}_j\,\omega_1+\omega_{j+1}\;=\;\varrho\,\omega_j,\qquad j=1,\dots,D .
\end{equation}
\end{Lemma}

\begin{proof}
(i) On \((0,1]\) the map \(\varphi\) is continuous and strictly decreasing (each summand is), \(\varphi(1)=R_++D\eta<1\) by the choice of \(\eta\), and \(\varphi(\varrho)\ge\tilde f^{\,+}_1\varrho^{-1}\to\infty\) as \(\varrho\downarrow0\) because \(\tilde f^{\,+}_1\ge\eta>0\). The intermediate value theorem gives a root in \((0,1)\), and strict monotonicity its uniqueness.
(ii) First, \(\omega_1=\varrho^{0}\sum_{i=1}^{D}\tilde f^{\,+}_i\varrho^{-i}=\varphi(\varrho)=1\). For \(1\le j\le D\),
\[
\varrho \omega_j=\varrho^{\,j}\sum_{i=j}^{D}\tilde f^{\,+}_i\varrho^{-i}
=\tilde f^{\,+}_j+\varrho^{\,j}\!\!\sum_{i=j+1}^{D}\!\!\tilde f^{\,+}_i\varrho^{-i}
=\tilde f^{\,+}_j\,\omega_1+\omega_{j+1},
\]
the last sum being empty for \(j=D\), consistent with \(\omega_{D+1}=0\). Finally \(\omega_j\ge\varrho^{\,j-1}\,\tilde f^{\,+}_j\varrho^{-j}=\tilde f^{\,+}_j/\varrho\).
\end{proof}

\begin{Remark}
Equation~\eqref{eq:app-weights} says that \(\omega\) is a left Perron eigenvector, with eigenvalue \(\varrho\), of the companion matrix of the inflated coefficients \(\tilde f^{\,+}\); the \(\eta\)-inflation only ensures strict positivity of \(\omega\) when some \(f^{+}_d=0\). The resulting \(\varrho\) depends on the chosen \(\eta\); it is a valid drift coefficient for the bounds below rather than an intrinsic rate, and can be taken arbitrarily close to the \(\eta\downarrow0\) root of \(\sum_d f^{+}_d\varrho^{-d}=1\).
\end{Remark}

\begin{TheoremMain}[geometric ergodicity]
\ThmGeoErgBody
\end{TheoremMain}

\begin{proof}
Throughout fix \(\eta:=(1-R_+)/(2D)\), let \(\varrho\in(0,1)\) and \(\omega=(\omega_1,\dots,\omega_D)\) be the root and weights of Lemma~\ref{lem:weights} for the inflated coefficients \(\tilde f^{\,+}_d:=f^{+}_d+\eta\), and set \(V(x):=1+\sum_{d=1}^{D}\omega_dx_d\). Every \(\omega_d>0\) and \((\varrho,\omega)\) depends only on \(f\), so \(V\) is of the form claimed in (i).

\emph{Step 1 (Markov structure).}
Under~\eqref{eq:app-model} the conditional law of \(N(t)\) given \(\mathcal{H}(t-1)\) depends on the past only through \(X(t-1)\), so \(X(t)\) is a time-homogeneous Markov chain on \(\mathbb{N}_0^{D}\): from \(x=(x_1,\dots,x_D)\) it moves to \((N',x_1,\dots,x_{D-1})\) with \(N'\sim\mathrm{NB}\{h_s(b+\sum_{d}f(d)x_d),\kappa\}\).

\emph{Step 2 (conditional-mean domination).}
For every \(x\in\mathbb{N}_0^{D}\), using the mean parameterization, Lemma~\ref{lem:link-envelope}, the subadditivity \((u+v)_+\le u_++v_+\) of the positive part, \(f(d)x_d\le f^{+}_dx_d\) for \(x_d\ge0\), and \(f^{+}_d\le\tilde f^{\,+}_d\),
\[
\mathbb{E}\{N'\mid x\}
= h_s\Bigl(b+\sum_{d}f(d)x_d\Bigr)
\;\le\;\Bigl(b+\sum_d f(d)x_d\Bigr)_{\!+}+s\log2+\epsilon
\;\le\;c_b+\sum_{d=1}^{D}\tilde f^{\,+}_d x_d,
\]
with \(c_b:=b_++s\log2+\epsilon<\infty\).

\emph{Step 3 (Foster--Lyapunov drift).}
By Step~1 the last \(D-1\) coordinates shift deterministically, so with Step~2 and then the weight identity~\eqref{eq:app-weights} (recall \(\omega_1=1\), \(\omega_{D+1}=0\)),
\[
\begin{aligned}
\mathbb{E}\{V(X(t))\mid X(t-1)=x\}
&=1+\omega_1\,\mathbb{E}\{N'\mid x\}+\sum_{j=1}^{D-1}\omega_{j+1}x_j\\
&\le1+c_b+\sum_{j=1}^{D}\bigl(\tilde f^{\,+}_j \omega_1+\omega_{j+1}\bigr)x_j
=\varrho V(x)+C,
\end{aligned}
\]
with \(C:=1-\varrho+c_b\). Fix \(\varrho'\in(\varrho,1)\) and put \(C_R:=\{x:V(x)\le C/(\varrho'-\varrho)\}\); on the complement of \(C_R\) we have \(\varrho V+C\le\varrho'V\), so
\[
PV\;\le\;\varrho'V+C\,\mathbf 1_{C_R},
\]
the geometric drift condition (V4), and \(C_R\) is a \emph{finite} set because every \(\omega_j>0\).

\emph{Step 4 (irreducibility, aperiodicity, small sets).}
For every state the intensity \(h_s(\cdot)\) is finite and strictly positive (\(1+e^{x/s}>1\)), so the negative-binomial law charges every point of \(\mathbb{N}_0\). Hence for any \(x,y\in\mathbb{N}_0^{D}\), prescribing the next \(D\) counts, in time order \(y_D,\dots,y_1\) so that after \(D\) steps the window reads \(y=(y_1,\dots,y_D)\), gives \(P^{D}(x,\{y\})>0\): the chain is irreducible with respect to counting measure. From the zero window \(\mathbf 0:=(0,\dots,0)\), \(P(\mathbf 0,\{\mathbf 0\})=\{\kappa/(\kappa+h_s(b))\}^{\kappa}>0\), so the common period is one and the chain is aperiodic. Finally, \(\beta_0:=\min_{x\in C_R}P^{D}(x,\{\mathbf 0\})>0\) as a minimum of finitely many positive numbers, so \(P^{D}(x,\cdot)\ge\beta_0\,\delta_{\mathbf 0}(\cdot)\) for all \(x\in C_R\): the finite set \(C_R\) is \(\nu_D\)-small (hence petite) with \(\nu_D=\beta_0\,\delta_{\mathbf 0}\).

\emph{Step 5 (conclusion).}
The chain is \(\psi\)-irreducible and aperiodic, and by Steps~3--4 it satisfies the drift condition (V4) towards a small set with \(V\ge1\) everywhere finite. The geometric ergodic theorem \citep[Theorem~15.0.1]{meyn2009markov} then gives positive Harris recurrence, a unique invariant probability \(\pi\) with \(\pi(V)<\infty\), and \(V\)-geometric ergodicity, which is the bound in (i). For (ii): under \(\pi\) all window coordinates share the marginal mean \(m:=\mathbb{E}_\pi\{N(t)\}\le\pi(V)<\infty\). Taking expectations of the conditional mean under stationarity and applying Lemma~\ref{lem:link-envelope}, the subadditivity of the positive part, and \(f(d)\le f^{+}_d\),
\[
m=\mathbb{E}_\pi\Bigl[h_s\Bigl(b+\sum_{d=1}^{D}f(d)\,N(t-d)\Bigr)\Bigr]
\;\le\;b_++s\log2+\epsilon+\sum_{d=1}^{D}f^{+}_d\,m
\;=\;c_b+R_+\,m;
\]
since \(m<\infty\) and \(R_+<1\), solving gives \(m\le c_b/(1-R_+)\), the bound in (ii).
\end{proof}

\begin{PropositionMain}[existence and input-tracking growth]
\PropGrowthBody
\end{PropositionMain}

\begin{proof}
For (i): every conditional law \(\mathrm{NB}\{\lambda(t),\kappa\}\) is a proper distribution on \(\mathbb{N}_0\), since \(h_s\) is finite and strictly positive on \(\mathbb{R}\), so \(\lambda(t)\in(0,\infty)\) whenever the current window is finite. Given the initial law of \(X(0)\), the Ionescu--Tulcea extension theorem \citep[see, e.g.,][]{kallenberg2002foundations} yields a process with exactly these conditionals, unique in law, and induction over \(t\) gives \(\Pr\{N(t)<\infty\text{ for all }t\}=1\): in discrete time there is no explosion phenomenon, and the substantive stability question is growth. For (ii): let \(\eta,\varrho,\omega,V\) be as fixed at the start of the proof of Theorem~\ref*{thm:gpdhp-stability}, so \(\varrho\) and \(\omega\) depend only on \(f\), and set \(c_b(t):=b(t)_++s\log2+\epsilon\). Steps~2--3 of that proof at time \(t\) involve the baseline only through \(b(t)_+\), giving \(\mathbb{E}\{V(X(t))\mid X(t-1)\}\le\varrho\,V(X(t-1))+C_t\) with \(C_t:=1-\varrho+c_b(t)\); iterating,
\(\mathbb{E}\{V(X(t))\}\le\varrho^{t}\,\mathbb{E}\{V(X(0))\}+\sum_{r=1}^{t}\varrho^{\,t-r}C_r\le\varrho^{t}\,\mathbb{E}\{V(X(0))\}+\max_{r\le t}C_r/(1-\varrho)\).
Since \(\omega_1=1\) gives \(N(t)\le V(X(t))\), and \(V(X(0))\le(1+\max_j\omega_j)\bigl\{1+\sum_{d=1}^{D}X_d(0)\bigr\}\), the displayed bound follows with \(C_f:=1+\max_j\omega_j\), using \(\max_{r\le t}C_r/(1-\varrho)=1+\bigl\{s\log2+\epsilon+\max_{r\le t}b(r)_+\bigr\}/(1-\varrho)\).
\end{proof}

\begin{Remark}[certificates for fitted baselines]
We deliberately attach no geometric-ergodicity certificate to individual fitted series: a fitted trend scale is never exactly zero (the selection searches a log-scale box with a positive lower bound), so no fitted baseline is exactly constant or periodic, and Theorem~\ref*{thm:gpdhp-stability} plays the role of the idealized benchmark. The certificate a fit earns from \(R_+<1\) is the input-tracking bound of Proposition~\ref*{prop:growth}: means uniformly bounded whenever its baseline path is bounded above (a nonpositive fitted trend slope, or bounded covariates) and at most linear mean growth otherwise; existence at all times holds for every kernel and is not part of the certificate.
\end{Remark}

\begin{Remark}[scope and sharpness]
(i) The proof uses only the conditional mean of the observation law and the fact that it charges every point of \(\mathbb{N}_0\), so the guarantee holds for every dispersion \(\kappa\in(0,\infty)\) and for Poisson observations; the Poisson--rectifier, constant-baseline case, for which \citet[Theorem~2.1]{costa2024general} establish the same sufficiency, is recovered in the \(\kappa\to\infty\), \(s\to0\), \(\epsilon=0\) limit.
(ii) The condition cannot be weakened to \(R_+\le1\): take \(f\ge0\) with \(\sum_df(d)=1\) and constant \(b>0\). If a stationary distribution with finite mean \(m\) existed then, since \(h_s(x)\ge x\) (Lemma~\ref{lem:link-envelope}), stationarity would give \(m\ge b+\sum_df(d)\,m=b+m\), a contradiction. Moreover, from any initial window with finite means, \(m_t:=\mathbb{E}\{N(t)\}\) satisfies \(m_t\ge b+\min_{1\le d\le D}m_{t-d}\), so the running window minimum gains at least \(b\) every \(D\) steps and \(m_t\to\infty\).
(iii) Only the positive part \(f^{+}_d=\{f(d)\}_+\) enters, so inhibitory lags can only help; the condition is sufficient but not necessary: for memory length two, \citet{costa2024stability} characterize the exact stability region, which strictly exceeds \(\{R_+<1\}\) in the presence of inhibition.
(iv) The threshold parallels the classical condition ``sum of feedback coefficients \(<1\)'' for linear Poisson autoregressions \citep{fokianos2009poisson}; here only the excitatory mass must satisfy it, at the price of the additive constant \(s\log2+\epsilon\) contributed by the link.
\end{Remark}

\phantomsection
\section*{Appendix C: Implementation details}\label{app:alfv}

This appendix collects implementation details: the fast evaluation of the excitation, and the machinery behind the stability-constrained fit of
Section~\ref*{sec:hyperparameter-selection}. Fix a hyperparameter vector \(\psi\) and write
\(R_+(\theta,\psi)=\sum_d\{m_{\mathrm{NB}}(\psi)_d+(L(\psi)\theta_g)_d\}_+\) for the positive
excitation mass of the fitted kernel. The constrained problem is the bilevel program
\[
\min_{\psi}\;G(\psi):=-\mathrm{FV}(\psi),\qquad
\theta^*(\psi)=\arg\min_{\theta}\;F_{\mathrm{tr}}(\theta,\psi)\ \ \text{s.t.}\ \ R_+(\theta,\psi)\le c,
\]
with \(c=1-\delta\), \(F_{\mathrm{tr}}\) the collapsed train MAP objective of
Section~\ref*{Inference}, and \(\mathrm{FV}\) the forward-validation criterion of
Equation~(\ref*{eq:fv-criterion}), evaluated at the constrained inner fit \(\theta^*(\psi)\).

\paragraph{Inner problem.} The constraint is imposed by an augmented Lagrangian: for a fixed
increasing sequence \(\rho_1<\dots<\rho_K\) (we use \(30\) to \(3\times10^{5}\)) and multiplier
\(\gamma\) initialised at zero, each subproblem
\[
\min_\theta\; F_{\mathrm{tr}}(\theta,\psi)+\tfrac{1}{2\rho_k}\Bigl[\bigl\{\max(0,\gamma+\rho_k\,(R_+-c))\bigr\}^2-\gamma^2\Bigr]
\]
is an unconstrained, piecewise-smooth MAP solved by the same L-BFGS used everywhere else (we adopt
the convention \(d(x_+)/dx|_{0}=0\) at the measure-zero lag- and augmented-branch crossings, and no iterate
landed exactly on such a kink in our runs), after which
the incoming multiplier is updated, \(\gamma\leftarrow\max\{0,\gamma+\rho_k(R_+-c)\}\). Within a fixed excitation-sign region and a fixed hinge branch, \(R_+\) is affine in
\(\theta\) and has zero Hessian, so the subproblem Hessian is \(H+a_k\rho_k gg^\top\), where
\(a_k=\mathbf 1\{\gamma+\rho_k(R_+-c)>0\}\) is the hinge-branch indicator,
\(H=\nabla^2_\theta F_{\mathrm{tr}}\), and \(g=\nabla_\theta R_+\) (a subgradient over the positive
lags). The multiplier is what pins the fit to the constraint surface: at a binding solution the
effective multiplier is positive and \(R_+\approx c\).

\paragraph{Outer hypergradient.} Write the final (\(k=K\)) augmented Lagrangian as the named function
\[
\mathcal{L}_{\rho_K,\gamma}(\theta,\psi)=F_{\mathrm{tr}}(\theta,\psi)+\tfrac{1}{2\rho_K}\bigl[\{\gamma+\rho_K(R_+-c)\}_+^2-\gamma^2\bigr],
\]
with \(\gamma\) the multiplier \emph{entering} this subproblem (fixed) and \(\theta^*\) its minimizer.
Fix the active set: the sign pattern of the excitatory lags that defines \(R_+\), together with the
branch of the hinge at \(\theta^*\), recorded by the indicator
\(a:=\mathbf 1\{\gamma+\rho_K(R_+(\theta^*)-c)>0\}\) (\(a=1\): cap active; \(a=0\): slack). On this set \(\mathcal{L}_{\rho_K,\gamma}\) is \(C^2\),
\(\theta^*\) solves \(\nabla_\theta\mathcal{L}_{\rho_K,\gamma}=0\), and the implicit-function theorem
applied to this smooth stationarity condition gives
\[
\boldsymbol\nu=\bigl[\nabla^2_{\theta\theta}\mathcal{L}_{\rho_K,\gamma}\bigr]^{-1}\partial_\theta\tilde G,
\qquad
\frac{d\tilde G}{d\psi}=\partial_\psi\tilde G-\boldsymbol\nu^\top\partial_\psi\nabla_\theta\mathcal{L}_{\rho_K,\gamma}.
\]
Because \(R_+\) is affine in \(\theta\) at a fixed sign pattern, and the hinge contributes curvature only on its active branch, \(\nabla^2_{\theta\theta}\mathcal{L}_{\rho_K,\gamma}=H+a\rho_Kgg^\top\)
with \(H=\nabla^2_\theta F_{\mathrm{tr}}\), \(g=\nabla_\theta R_+\), and
\(\partial_\psi\nabla_\theta\mathcal{L}_{\rho_K,\gamma}=\partial_\psi\nabla_\theta F_{\mathrm{tr}}+\gamma^*_{\mathrm{eff}}\,\partial_\psi g+a\rho_K\,g\,\partial_\psi R_+\),
where \(\gamma^*_{\mathrm{eff}}=\{\gamma+\rho_K(R_+(\theta^*)-c)\}_+=a\,\{\gamma+\rho_K(R_+(\theta^*)-c)\}\) is the \emph{effective} multiplier at \(\theta^*\)
(the quantity the \(\gamma\)-update converges to), and \(\partial_\psi g=\partial_\psi\nabla_\theta R_+\) is the Jacobian of \(g\) in \(\psi\) at fixed \(\theta\). In this differentiation only the incoming multiplier \(\gamma\) and the active branch are held fixed: \(\gamma^*_{\mathrm{eff}}\) itself varies with \((\theta,\psi)\) through \(R_+\), and that variation is precisely what produces the terms proportional to \(a\rho_K\) here and in the Hessian. The direct term \(\partial_\psi R_+\) collects the dependence of the positive mass on \(\psi\) through \(m_{\mathrm{NB}}\), \(L\), and the other kernel hyperparameters. Sherman--Morrison on the two solves
\(\boldsymbol\nu_0=H^{-1}\partial_\theta\tilde G\) and \(h_g=H^{-1}g\) then gives
\[
\boldsymbol\nu=\boldsymbol\nu_0-\frac{a\rho_K\,(g^\top\boldsymbol\nu_0)}{1+a\rho_K\,(g^\top h_g)}\,h_g,
\qquad
\frac{d\tilde G}{d\psi}=\partial_\psi\tilde G-\boldsymbol\nu^\top\bigl(\partial_\psi\nabla_\theta F_{\mathrm{tr}}+\gamma^*_{\mathrm{eff}}\,\partial_\psi g\bigr)-a\rho_K(\boldsymbol\nu^\top g)\,\partial_\psi R_+ .
\]
This is exact for the final augmented subproblem at the fixed active set; it is not a derivative
of the exact inequality-constrained solution map, which is nondifferentiable where the active set
changes (discussed below). On the active branch the cost is two linear solves with \(H\)
(\(\boldsymbol\nu_0\) and \(h_g\)), one more than the unconstrained
adjoint of Equation~(\ref*{eq:fv-hypergrad}); on the slack branch \(h_g\) is not needed and the
cost is the single unconstrained solve. The denominator exceeds one whenever \(H\) is positive
definite at \(\theta^*\), verified numerically below, so the update is well conditioned. When the cap
is slack, \(a=0\) and \(\gamma^*_{\mathrm{eff}}=0\), so every \(\rho_K\)-term vanishes and the formula
reduces exactly to the unconstrained adjoint of Equation~(\ref*{eq:fv-hypergrad}); our implementation
gates the \(\rho_K\)-terms on exactly this indicator (at a slack solution the analytic and
unrolled-Newton gradients agree to \(\sim10^{-8}\); see the verification below).

\paragraph{Feasibility penalty.} The validation loss alone cannot see a constraint violation:
\(G\) is evaluated \emph{at} \(\theta^*\), so a \(\psi\) whose inner fit fails to reach the budget
still scores well and can be selected. Such \(\psi\) exist because
\(R_+(m_{\mathrm{NB}})=K_{\mathrm{NB}}\) exactly (the parametric kernel is normalised), while a
Gaussian-process amplitude \(\sigma_g\) near zero leaves \(\theta_g\) too weak to compensate against
the unit ridge. We therefore optimise the penalised outer objective
\[
\tilde G(\psi)=G(\psi)+\mu\bigl\{\max(0,R_+(\theta^*(\psi),\psi)-c)\bigr\}^2,\qquad \mu=10^{5},
\]
whose gradient is obtained by the same adjoint. The penalty is identically zero at every feasible
\(\psi\), so it never perturbs a cell whose fit already satisfies the cap, and elsewhere supplies
the gradient that steers the multi-start out of the infeasible region. This is where the finite-\(\rho\)
adjoint matters: the \(\rho_K\to\infty\) (bordered) limit enforces \(dR_+/d\psi=0\) and would
annihilate precisely this term.

\paragraph{Search settings.} The outer search samples \(20\) starting points over the box of
admissible hyper\-parameters (uniformly for bounded coordinates, log-uniformly for scale and
variance coordinates) and refines each by a bound-constrained L-BFGS with a Mor\'e--Thuente line
search on the penalised hypergradient (\(600\) outer steps; \(1{,}000\) inner steps per \(\rho\) of
the ramp above; \(\delta=10^{-4}\)); the starts run concurrently over a CPU thread pool and the
best validation score is kept. The multi-start matters: a single local descent on the higher-count
daily series readily settles into inferior optima, for example a weak weekly-baseline solution that
leaves day-of-week structure in the excitation. The discrete choices of Section~\ref*{sec:hyperparameter-selection}
are handled by repeating the search per annual-harmonic count. The whole search runs under a fixed
random seed and completes in one to three minutes per fit on a multicore laptop CPU.

\paragraph{Local regularity and verification.} The implicit-function steps above require
\(\theta^*(\psi)\) to be an isolated stationary point with positive-definite Hessian \(H\). This is a
genuine assumption: \(F_{\mathrm{tr}}\) is smooth but not globally convex in \(\theta\): for a zero
count the negative-binomial term \(\kappa\log\{\kappa+\lambda(t)\}\) is concave in the mean, so the
likelihood can contribute negative curvature that the unit whitened ridge is not guaranteed to
dominate everywhere. We therefore report it as an implementation diagnostic rather than assert it: at every
reported fit (the selected inner optimum and the full-period refit, across all seven cells) the
smallest eigenvalue of the exact inner Hessian \(H\) is numerically at least \(1.0\): the ridge contributes
exactly one to every eigenvalue, and at the fitted solutions the likelihood curvature did not push
any eigenvalue measurably below that floor (smallest value observed \(1.0000000\) to seven digits, at a fit with a prior block
switched off; largest such minimum \(1.00003\)). The design \(A=[\,B\mid XL\,]\), with all prior scale factors folded into \(B\) and \(L\), likewise had full column rank on each cell's full fitting-period design at the selected hyperparameters; the smallest singular value over the seven final selected fits (not over the inner optimization iterates) was \(1.4\times10^{-2}\). A second nonsmoothness deserves note: \(R_+\)
is piecewise linear, \(g\) is a subgradient, and \(\theta^*(\psi)\) is in general nondifferentiable
at hyperparameter values where the active set of positive lags changes; the finite-\(\rho_K\)
subproblem is \(C^2\) within a fixed active set and serves as the practical smoothing of this kink. The hypergradient was
verified against automatic differentiation through an unrolled Newton solve of the same subproblem
at a fixed active set, in three regimes (cap slack, cap binding, penalty active). When the cap is
slack the agreement is exact, at relative accuracy of order \(10^{-8}\), the unconstrained adjoint.
When the cap binds or the penalty is active, every coordinate that does not reshape the fitted
kernel agrees to \(10^{-4}\)--\(10^{-2}\), while the kernel coordinates can differ by order one:
at such solutions the multiplier pins lags of the fitted kernel exactly at zero, so those
derivatives are one-sided (the active-set nonsmoothness just described) and any
subgradient choice is legitimate there. Because the
augmented Lagrangian satisfies the constraint only in the limit \(\rho\to\infty\), the achieved
\(R_+\) can exceed the nominal budget by \(O(10^{-5})\); we therefore verify and report the
\emph{achieved} \(R_+\) for every fit, which is the quantity
Theorem~\ref*{thm:gpdhp-stability} requires to be below one, rather than the requested budget.

\paragraph{Fast excitation evaluation.} We implement the excitation itself by fast convolution. The one-step excitation \(\sum_{d\ge1} f(d)\,N(t-d)\), with length-\(D_{\max}\) kernel \(f=m_{\mathrm{NB}}+L\theta_g\), is a causal convolution of the count sequence; we evaluate it and its adjoint (the latter needed for the gradient and the analytic bilevel hypergradient) by the real FFT in \(O(T\log T)\) time, in place of the direct \(O(TD_{\max})\) summation (and without ever forming the \(T\times D_{\max}\) lagged-count design or its \(O(TD_{\max}^2)\) product with \(L\)). Only the kernel matrix--vector product \(L\theta_g\) then remains, at \(O(D_{\max}^2)\), so a single likelihood, gradient, or hypergradient evaluation costs \(O(T\log T+D_{\max}^2)\). This is an exact reindexing of the same finite sum (it matches the direct evaluation to machine precision, not an approximation) and is the default throughout our implementation: MAP fitting, the forward-validation selection, and the projection. It is essentially break-even at the modest horizon \(D_{\max}=100\) used in our experiments and increasingly advantageous as the horizon grows: on the daily worldwide-terrorism series (\(T\approx1.3\times10^{4}\)) a single bilevel evaluation (inner fit plus hypergradient) runs about three times faster at \(D_{\max}=400\), the regime in which the direct form's per-evaluation \(O(TD_{\max})\) cost and \(O(TD_{\max}^2)\) design build would otherwise dominate.

\paragraph{Structured covariance products.} The baseline and lag covariances entering the collapsed objective admit fast matrix--vector products. From Equation~(\ref*{eq:Kb-fourier}),
\[
K_b^{\mathrm{num}}v = \sigma_{\mathrm{level}}^2\mathbf{1}(\mathbf{1}^\top v)+\sigma_{\mathrm{lin}}^2\mathbf{t}(\mathbf{t}^\top v)+\sigma_{\mathrm{season}}^2 H_K(H_K^\top v)+\sigma_{\mathrm{week}}^2 H_{K_w}^{(P_w)}\bigl(H_{K_w}^{(P_w)\top} v\bigr)+\epsilon_b^2 v,
\]
so the level and trend terms are rank-one and the seasonal term costs \(O(TK)\) for \(K\) harmonics. The lag covariance factorizes as \(K_g=D_a K_{\mathrm{stat}}D_a\) with \(D_a=\operatorname{diag}\{a(1),\ldots,a(D_{\max})\}\) and \(K_{\mathrm{stat}}\) the RBF covariance on the warped inputs \(w(d)\); when \(D_{\max}\) is large, structured-kernel interpolation on a uniform grid in the warped coordinate gives products with \(K_{\mathrm{stat}}\) in \(O(M\log M)\) time, with interpolation cost linear in \(D_{\max}\).

\phantomsection
\section*{Appendix D: Benchmark specifications and evaluation protocol}\label{app:benchmarks}

The parametric Hawkes benchmarks use the negative-binomial excitation kernel family of Section~\ref*{sec:model} and differ only in their baseline intensity \(\mu(t)\): Discrete DHP, \(\mu(t)=\gamma_0\); Linear DHP, \(\mu(t)=\gamma_0+\gamma_1 t\); Sinusoidal DHP, \(\mu(t)=\gamma_0+\gamma_1\sin(2\pi t/P)\); and Linear\,+\,Sinusoidal DHP, \(\mu(t)=\gamma_0+\gamma_1 t+\gamma_2\sin(2\pi t/P)\), with \(P=52\) for the weekly series and \(P=365\) for the daily series. The baseline-only benchmark removes the lagged Hawkes term and uses the same finite-Fourier baseline class as GP-DHP. The NB-INGARCH benchmark is a negative-binomial count model whose conditional mean is driven by lagged counts and lagged conditional means, reimplemented in the same JAX framework and validated against the R \texttt{tscount} package \citep{fokianos2009poisson,liboschik2017jss}; for the two daily shooting series an NB-INGARCH+cov variant receives the same temperature-anomaly and holiday covariates as GP-DHP, with scales selected under the same forward-validation protocol. The Histogram DHP-NB uses nonnegative fixed-bin excitation coefficients: for fixed lag bins \(B_m\) with \(S_m(t)=\sum_{d\in B_m}N(t-d)\), it fits \(\zeta(t)=\mu(t)+\sum_m\theta_m S_m(t)\), \(\theta_m\ge0\), under the same rectified negative-binomial observation layer, with the \(L_2\) penalty on \(\theta\) its single tuned hyperparameter (forward-validated on the same last-40\% window as GP-DHP); it is related to histogram and basis-expanded Hawkes kernels \citep{lewis2011,zhou2013}. The Random-histogram DHP-NB of \citet{browning2022} instead treats the histogram structure as random, placing a prior on the number and location of the excitation breakpoints and inferring them by reversible-jump MCMC \citep{green1995reversible} under the same observation, baseline, and covariate design, so the two histogram models differ only in whether the bin structure is prespecified and optimized or random and sampled; its one-step predictive distribution is the posterior mixture over the sampled kernels, evaluated by log-sum-exp averaging. The discrete GP-Hawkes is the discrete-time analogue of the Gaussian-process--modulated Hawkes processes of \citet{zhou2020} and \citet{zhang2020}: over the lag grid it uses a squared-GP kernel \(g(d)^2\) with \(g\sim\mathcal{GP}(0,\sigma^2 K_{\mathrm{RBF}})\)---a nonnegative, fully nonparametric kernel with no parametric backbone---together with a softplus intensity link matching that of GP-DHP, under the same negative-binomial observation and baseline/covariate design as the Histogram DHP-NB. All of its continuous hyperparameters (the Gaussian-process amplitude and length-scale, the softplus bandwidth, and the negative-binomial dispersion) are selected by the \emph{identical} forward-validation protocol used for GP-DHP (Section~\ref*{sec:hyperparameter-selection}), with the Gaussian-process coefficients fitted by MAP at each inner solve.

\paragraph{Predictive scoring and significance testing.} Each model is scored by one-step-ahead predictive log-likelihood on the held-out test period, \(\mathrm{pLL}=\sum_{t\in\mathcal{T}_{\mathrm{test}}}\log p(N(t)\mid\mathcal{H}(t-1))\), where \(\mathcal{H}(t-1)=\{N(s):s<t\}\) and all negative-binomial normalizing constants are included; these are plug-in scores conditional on the fitted coefficients and selected hyperparameters, with parameter and hyperparameter uncertainty not integrated out. Per-observation log-score sequences are compared against GP-DHP by a Diebold--Mariano statistic \citep{diebold1995comparing} with a Newey--West heteroskedasticity- and autocorrelation-consistent variance at bandwidth \(\lfloor n^{1/3}\rfloor\) and the small-sample correction of \citet{harvey1997testing}, giving a two-sided \(p\)-value for each gap; the family-wise error rate is controlled by a Holm correction applied within each fitted cell across its nine count-process comparisons (the four neural benchmarks of Table~\ref{tab:supp-neural} form a separate four-comparison family per cell). Because every model is selected once on the fitting period and then held fixed over the test period, these are conditional finite-test-sample comparisons of the fitted methods rather than tests of population-level predictive ability, and results for benchmarks nested or nearly nested in GP-DHP should be read with corresponding caution.

\begin{table}[H]
\centering
\caption{Selected settings for the additional count-process benchmarks. Histogram DHP-NB uses fixed lag bins and a fixed baseline form (both set by frequency), with an \(L_2\) penalty on the nonnegative bin coefficients as its only tuned hyperparameter; the penalty is chosen by the same forward-validation criterion as GP-DHP, on the last-40\% window of the fitting period. NB-INGARCH is a negative-binomial count time-series model with lagged conditional-mean and lagged-count terms, reimplemented in the same JAX framework and validated against the R \texttt{tscount} package; on the daily series it is also given the external temperature-anomaly and holiday covariates (NB-INGARCH+cov). The random-histogram DHP-NB is sampled by reversible-jump MCMC with four chains of \(4.8\times10^{4}\) sweeps; its posterior places from about two excitation bins on dengue to nine on the daily shooting series. All settings are for the unified all-JAX pipeline.}
\label{tab:supp-extra-benchmark-settings}
\small
\setlength{\tabcolsep}{5pt}
\resizebox{\textwidth}{!}{%
\begin{tabular}{lllllll}
\toprule
Dataset & Hist. bins & Hist. baseline & Hist. penalty (FV) & INGARCH obs. lags & Mean lag & INGARCH covariates \\
\midrule
Singapore dengue         & medium-log  & seasonal     & 1.72 & 1,\,4 & 1 & trend \\
German cryptosporidiosis & medium-log  & seasonal     & $10^{-8}$  & 1,\,4 & 1 & trend \\
NYC shootings            & equal-width & seasonal+DOW & $9\times10^{-8}$ & 1,\,7 & 1 & annual+DOW\,(+temp+hol.) \\
GVA shootings            & equal-width & seasonal+DOW & $3\times10^{-5}$ & 1,\,7 & 1 & annual+DOW\,(+temp+hol.) \\
Worldwide terrorism (RAND) & equal-width & seasonal+DOW & 0.0099 & 1,\,7 & 1 & annual+DOW \\
\bottomrule
\end{tabular}%
}
\end{table}

\paragraph{Note on neural benchmarks.} For completeness we also evaluated four neural autoregressive count models, negative-binomial multilayer-perceptron, GRU, LSTM, and DeepAR predictors \citep{rumelhart1986learning,cho2014learning,hochreiter1997long,salinas2020deepar}, under the same held-out protocol; their scores are reported in Table~\ref{tab:supp-neural} rather than in the main comparison. GP-DHP attains the highest score in every cell and beats all four by a Holm-significant margin on every series (the smallest gap is on NYC, about \(0.04\)--\(0.07\) nats per observation; the largest is on worldwide terrorism, where DeepAR-NB's one-step forecasts deteriorate on the bursty high-intensity held-out regime). Consistent with earlier experiments, these comparatively high-capacity predictors do not improve on the interpretable count-process benchmarks and appear sensitive to the limited training lengths of the shorter surveillance series; they also provide no baseline/excitation decomposition. As discussed in Section~\ref*{sec:real-data}, a neural point-process model such as the Neural Hawkes process \citep{mei2017} targets continuous-time event streams rather than the binned-count setting studied here. We therefore report the count-process comparisons as the primary evaluation.

\begin{table}[H]
\centering
\caption{Held-out one-step-ahead negative-binomial predictive log-likelihoods for the four neural autoregressive count models, on the same held-out periods and against the same GP-DHP reference (in bold) as the main comparison; each is tuned by a \(40\%\)-window forward-validation grid. Columns ``NYC$+$c'' and ``GVA$+$c'' are the covariate-augmented daily shooting fits. Superscripts give the Holm-adjusted Diebold--Mariano \(p\)-value against GP-DHP within each cell's four neural comparisons: \(^{*}p<0.05\), \(^{**}p<0.01\), \(^{***}p<0.001\). GP-DHP is highest in every cell and every neural model is significantly worse on every series.}
\label{tab:supp-neural}
\small
\setlength{\tabcolsep}{3.5pt}
\resizebox{\textwidth}{!}{%
\begin{tabular}{lrrrrrrr}
\toprule
 & \multicolumn{3}{c}{No external covariates} & \multicolumn{4}{c}{Daily shooting series} \\
\cmidrule(lr){2-4}\cmidrule(lr){5-8}
Model & Dengue & Crypto. & Terror. & NYC & NYC$+$c & GVA & GVA$+$c \\
\midrule
\textbf{GP-DHP} & \textbf{-1315.9} & \textbf{-947.3} & \textbf{-5079.5} & \textbf{-1965.7} & \textbf{-1949.3} & \textbf{-3055.9} & \textbf{-2985.0} \\
\midrule
MLP-NB    & $-1563.2^{***}$ & $-1101.6^{***}$ & $-7104.3^{***}$  & $-2015.5^{***}$ & $-2007.7^{***}$ & $-3206.3^{***}$ & $-3193.5^{***}$ \\
GRU-NB    & $-1547.4^{***}$ & $-1106.9^{***}$ & $-7663.3^{***}$  & $-2012.8^{***}$ & $-2006.0^{***}$ & $-3286.6^{***}$ & $-3212.0^{***}$ \\
LSTM-NB   & $-1543.7^{***}$ & $-1102.7^{***}$ & $-7699.1^{***}$  & $-2013.8^{***}$ & $-2026.7^{***}$ & $-3291.9^{***}$ & $-3261.6^{***}$ \\
DeepAR-NB & $-1473.8^{**}$  & $-991.9^{***}$  & $-11136.2^{***}$ & $-2008.4^{***}$ & $-2006.8^{***}$ & $-3316.1^{***}$ & $-3367.3^{***}$ \\
\bottomrule
\end{tabular}%
}
\end{table}

\phantomsection
\section*{Appendix E: Additional real-data results}\label{app:supp-results}

\begin{figure}[H]
\centering
\includegraphics[width=0.98\textwidth]{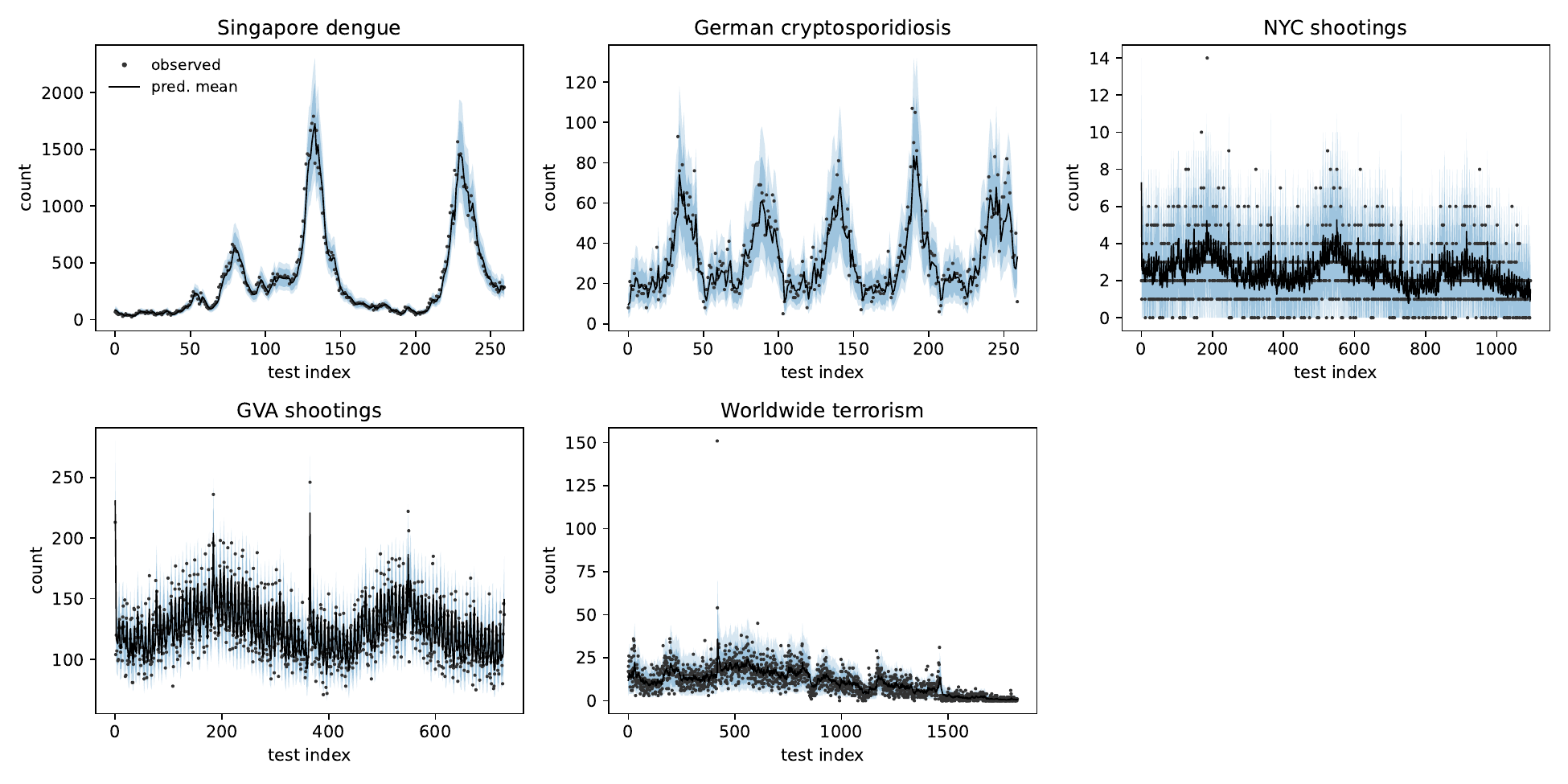}
\caption{Held-out one-step predictive intervals for the selected GP-DHP fits. Points show observed counts, black lines show one-step predictive means, and grey bands show central 80\% and 95\% negative-binomial predictive intervals. Each panel uses its own y-scale.}
\label{fig:supp-predictive-intervals}
\end{figure}

\begin{figure}[H]
\centering
\includegraphics[width=0.98\textwidth]{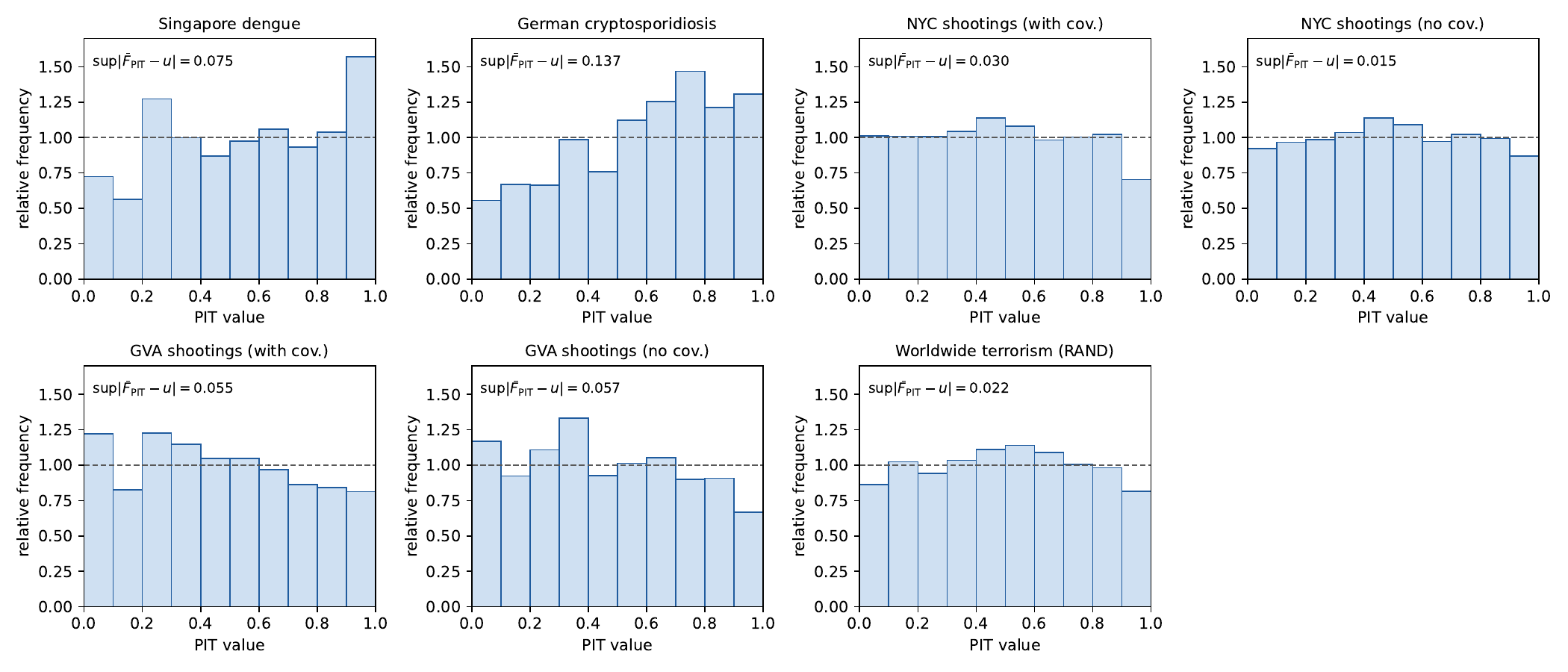}
\caption{Non-randomized probability-integral-transform histograms \citep{czado2009} for the held-out one-step negative-binomial predictives of the stability-constrained GP-DHP fits, one panel per fit: the five headline fits together with the no-covariate variants of the two daily shooting series (the same seven cells as the predictive tables). Each bar is the aggregated PIT mass in a decile bin (relative frequency, so the uniform reference, dashed, is \(1\)); the annotation is the supremum distance \(\sup_{u\in[0,1]}|\bar F_{\mathrm{PIT}}(u)-u|\) of the aggregated PIT CDF from the uniform CDF. Deviations below the uniform line in the top bins indicate conservative upper tails; the top-bin excess for Singapore dengue and the high-value tilt for German cryptosporidiosis are visible directly.}
\label{fig:supp-pit}
\end{figure}

This appendix reports, for the five real-data experiments, held-out error metrics for the GP-DHP point forecasts (Table~\ref{tab:supp-alt-metrics}), the forward-validation-selected model settings (Table~\ref{tab:supp-selected-hyperparameters}), excitation summaries for the fitted kernels (Table~\ref{tab:supp-excitation-summaries}), and the held-out one-step predictive intervals and probability-integral-transform histograms (Figures~\ref{fig:supp-predictive-intervals} and~\ref{fig:supp-pit}).

\begin{table}[H]
\centering
\caption{Held-out one-step mean absolute error (MAE) and root-mean-square error (RMSE) for the GP-DHP point forecasts, computed from the same stability-constrained fits as the main real-data comparison. Errors are on the count scale, so their magnitude reflects the typical count level of each series.}
\label{tab:supp-alt-metrics}
\small
\setlength{\tabcolsep}{8pt}
\begin{tabular}{lrr}
\toprule
Dataset & MAE & RMSE \\
\midrule
Singapore dengue & 45.786 & 80.323 \\
German cryptosporidiosis & 7.495 & 10.152 \\
NYC shootings & 1.238 & 1.583 \\
GVA shootings & 11.650 & 14.785 \\
Worldwide terrorism (RAND) & 3.593 & 5.757 \\
\bottomrule
\end{tabular}
\end{table}

\begin{table}[H]
\centering
\caption{Forward-validation-selected GP-DHP hyperparameters for the final real-data fits. Here $s$ is the softplus link scale, $K_{\mathrm{fou}}$ the number of annual harmonics, $\sigma_{\mathrm{level}}$, $\sigma_{\mathrm{season}}$, $\sigma_{\mathrm{week}}$, $\sigma_{\mathrm{lin}}$ the baseline level, annual, weekly, and linear-trend scales, $(K_{\mathrm{NB}},\text{mean lag},\text{NB size})$ the negative-binomial kernel mass and shape, and $(\beta,\sigma_g,\ell_g)$ the warped-RBF GP-correction parameters. A dash denotes a block that is absent (weekly for the weekly series) or inactive. For the two daily shooting series the external-covariate block additionally carries forward-validation-selected prior scales (temperature anomaly, holidays), $(0.06,\,1.29)$ for NYC and $(1.82,\,19.76)$ for GVA. These are the values selected under the stability constraint of Section~\ref*{sec:hyperparameter-selection}. Scale parameters printed as \(0.0000\) are at or near the bottom of their strictly positive search box (below \(5\times10^{-5}\)), not exactly zero.}
\label{tab:supp-selected-hyperparameters}
\small
\setlength{\tabcolsep}{3.2pt}
\resizebox{\textwidth}{!}{%
\begin{tabular}{lrrrrrrrrrrrr}
\toprule
Dataset & $\kappa$ & $s$ & $K_{\mathrm{fou}}$ & $\sigma_{\mathrm{level}}$ & $\sigma_{\mathrm{season}}$ & $\sigma_{\mathrm{week}}$ & $\sigma_{\mathrm{lin}}$ & $K_{\mathrm{NB}}$ & mean lag & NB size & $\beta$ & $\sigma_g$ / $\ell_g$ \\
\midrule
Singapore dengue          & 40.33  & 0.163 & 3 & 0.731 & 0.635  & --     & 0.0000 & 1.148 & 0.43  & 37.286 & 0.319 & 0.012 / 2.44  \\
German cryptosporidiosis  & 16.78  & 2.000 & 2 & 18.88 & 18.51  & --     & 0.0040 & 1.228 & 0.87  & 0.439  & 0.236 & 4.141 / 29.94 \\
NYC shootings             & 16.33  & 1.269 & 2 & 1.891 & 0.156  & 0.055  & 0.0000 & 1.209 & 39.29 & 0.710  & 0.064 & 0.003 / 1.07  \\
GVA shootings             & 148.71 & 0.052 & 3 & 2.455 & 13.529 & 17.257 & 0.0053 & 0.022 & 35.40 & 8.949  & 0.051 & 1.060 / 1.00  \\
Worldwide terrorism (RAND) & 6.88   & 0.462 & 2 & 3.748 & 0.010  & 10.880 & 0.0000 & 0.769 & 4.64  & 0.893  & 0.053 & 0.007 / 30.00 \\
\bottomrule
\end{tabular}%
}
\end{table}

\begin{table}[H]
\centering
\caption{Excitation summaries for the final GP-DHP real-data fits. Here \(R_+ = \sum_{d=1}^{D_{\max}} \max\{\widehat f(d),0\}\)  is the positive-mass diagnostic, and accumulation lags are computed from the positive part of the projected excitation.}
\label{tab:supp-excitation-summaries}
\small
\setlength{\tabcolsep}{5pt}
\begin{tabular}{lrrrrr}
\toprule
Dataset & $R_+$ & Peak lag & 50\% lag & 80\% lag & 90\% lag \\
\midrule
Singapore dengue          & 0.9999 & 1 & 1  & 2  & 2  \\
German cryptosporidiosis  & 0.6995 & 1 & 1  & 2  & 9  \\
NYC shootings             & 0.9999 & 1 & 17 & 48 & 68 \\
GVA shootings             & 0.9998 & 1 & 6  & 15 & 27 \\
Worldwide terrorism (RAND) & 0.9999 & 1 & 6  & 28 & 51 \\
\bottomrule
\end{tabular}

\vspace{0.25em}
\begin{minipage}{0.92\textwidth}
\footnotesize \(R_+\) sums only the positive (excitatory) part of the signed fitted lag response and is exactly the quantity in the sufficient stability condition \(R_+<1\), shown in \hyperref[app:stability]{Appendix~B} to guarantee stability of the negative-binomial--softplus GP-DHP itself, geometric ergodicity under a constant baseline and, in general, mean growth of at most the order of the running maximum of the positive part of the baseline (existence at all times holds without any condition), extending the Poisson--rectifier case of \citet{costa2024general} (Section~\ref*{sec:stability}). Every fit is selected under the constraint \(R_+\le1-\delta\) (\(\delta=10^{-4}\)) of Section~\ref*{sec:hyperparameter-selection}, so each satisfies \(R_+<1\), the excitation hypothesis of the stability results of \hyperref[app:stability]{Appendix~B}; the achieved values are reported here rather than the nominal budget, since the augmented Lagrangian meets the constraint only asymptotically. Cryptosporidiosis is the one series whose fit satisfies the condition with room to spare (\(R_+=0.70\)), so its constraint is inactive; on the other four the constraint binds and \(R_+\) sits at the budget.
\end{minipage}
\end{table}

\phantomsection
\section*{Appendix F: Search ranges and excitation-prior illustration}\label{app:supp-synthetic}

\begin{figure}[H]
\centering

\begin{subfigure}{0.32\linewidth}
  \centering
  \includegraphics[width=\linewidth]{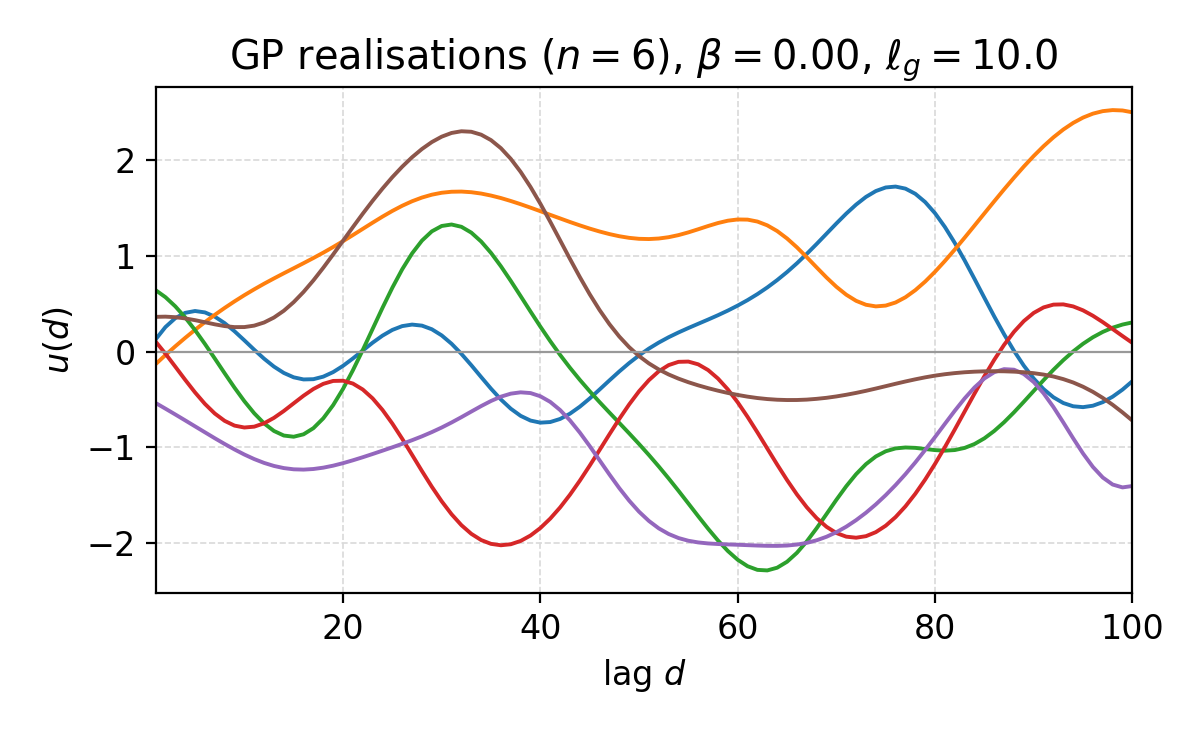}
  \caption{$u(d)$ draws, $\beta=0.00$, $\ell_g=10$}
\end{subfigure}\hfill
\begin{subfigure}{0.32\linewidth}
  \centering
  \includegraphics[width=\linewidth]{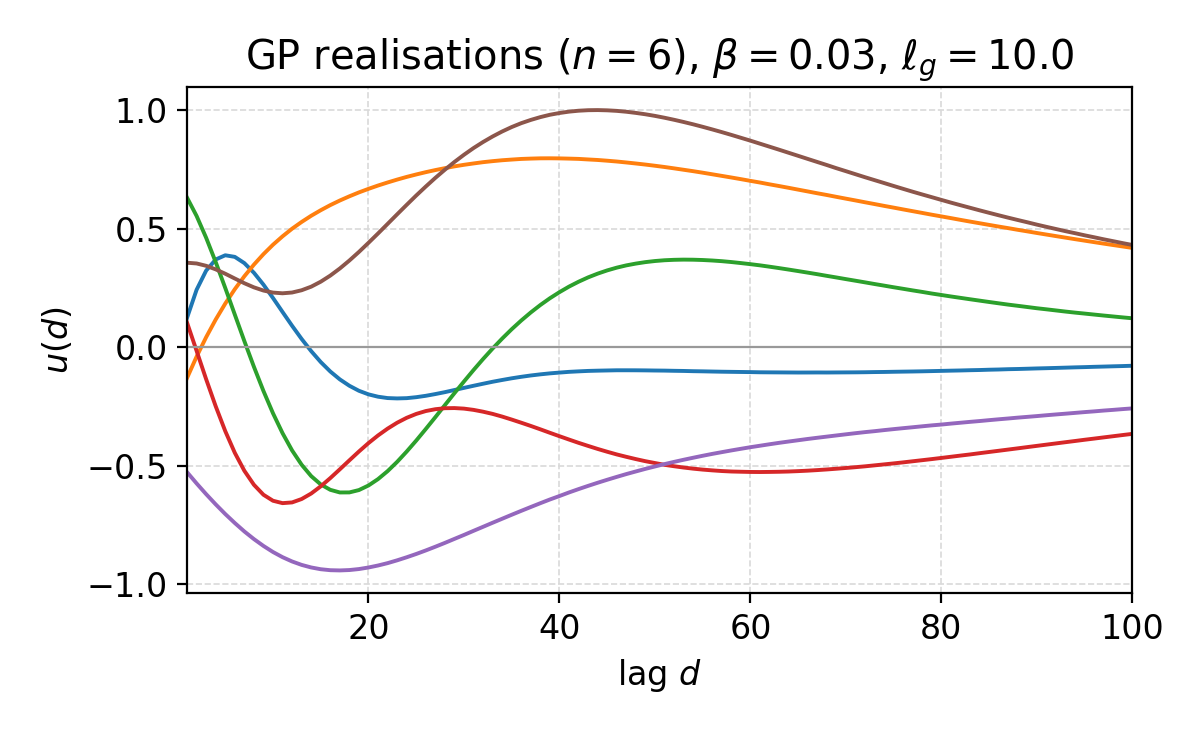}
  \caption{$u(d)$ draws, $\beta=0.03$, $\ell_g=10$}
\end{subfigure}\hfill
\begin{subfigure}{0.32\linewidth}
  \centering
  \includegraphics[width=\linewidth]{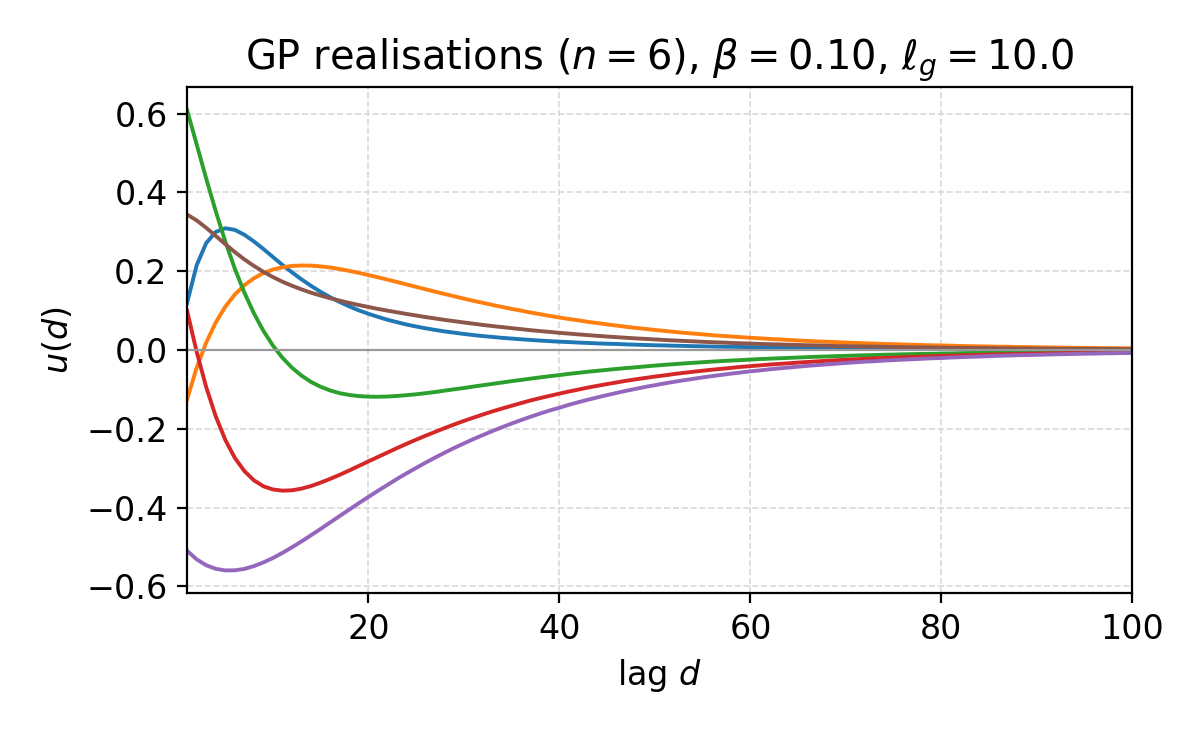}
  \caption{$u(d)$ draws, $\beta=0.10$, $\ell_g=10$}
\end{subfigure}

\vspace{0.6em}

\begin{subfigure}{0.32\linewidth}
  \centering
  \includegraphics[width=\linewidth]{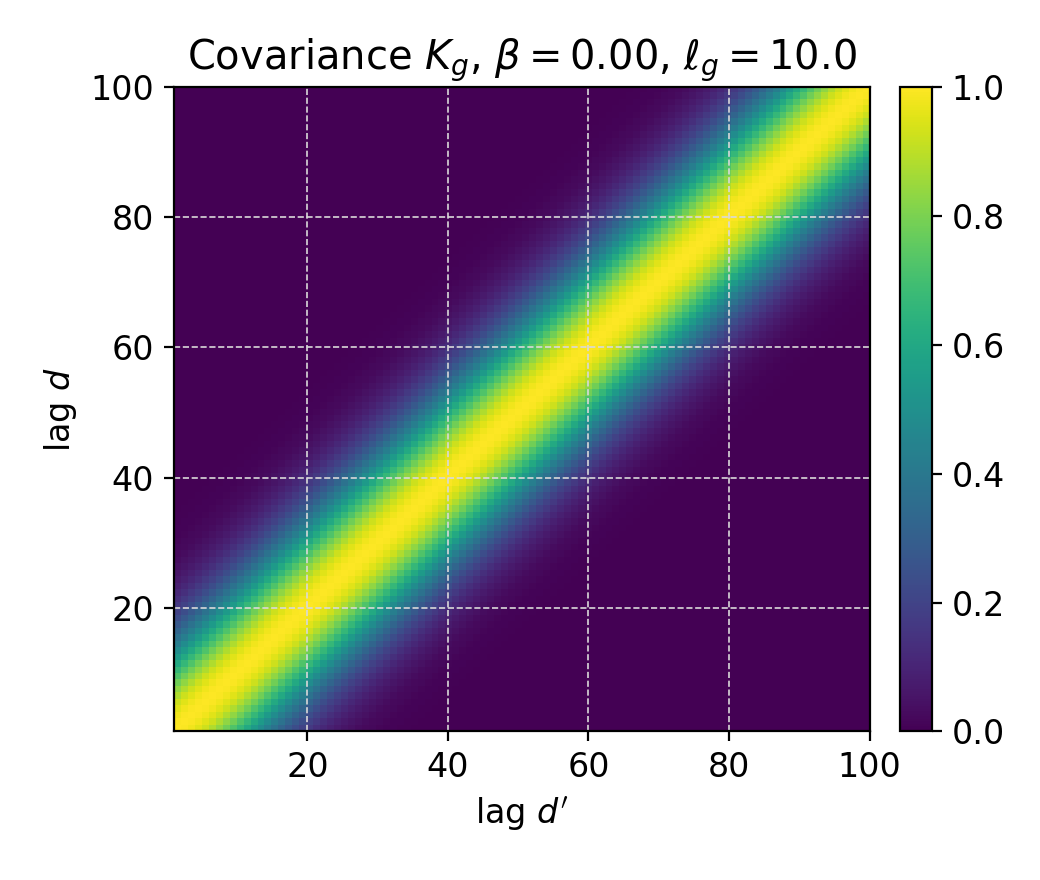}
  \caption{$K_g$ covariance heatmap, $\beta=0.00$, $\ell_g=10$}
\end{subfigure}\hfill
\begin{subfigure}{0.32\linewidth}
  \centering
  \includegraphics[width=\linewidth]{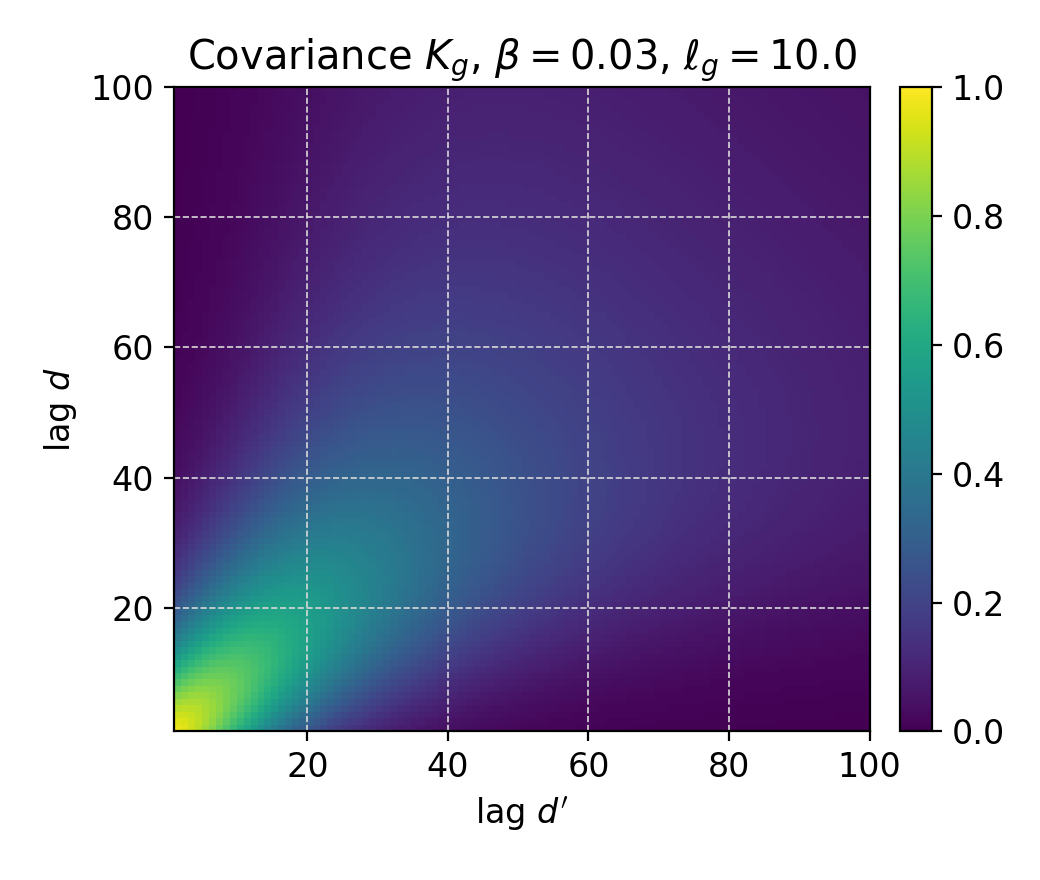}
  \caption{$K_g$ covariance heatmap, $\beta=0.03$, $\ell_g=10$}
\end{subfigure}\hfill
\begin{subfigure}{0.32\linewidth}
  \centering
  \includegraphics[width=\linewidth]{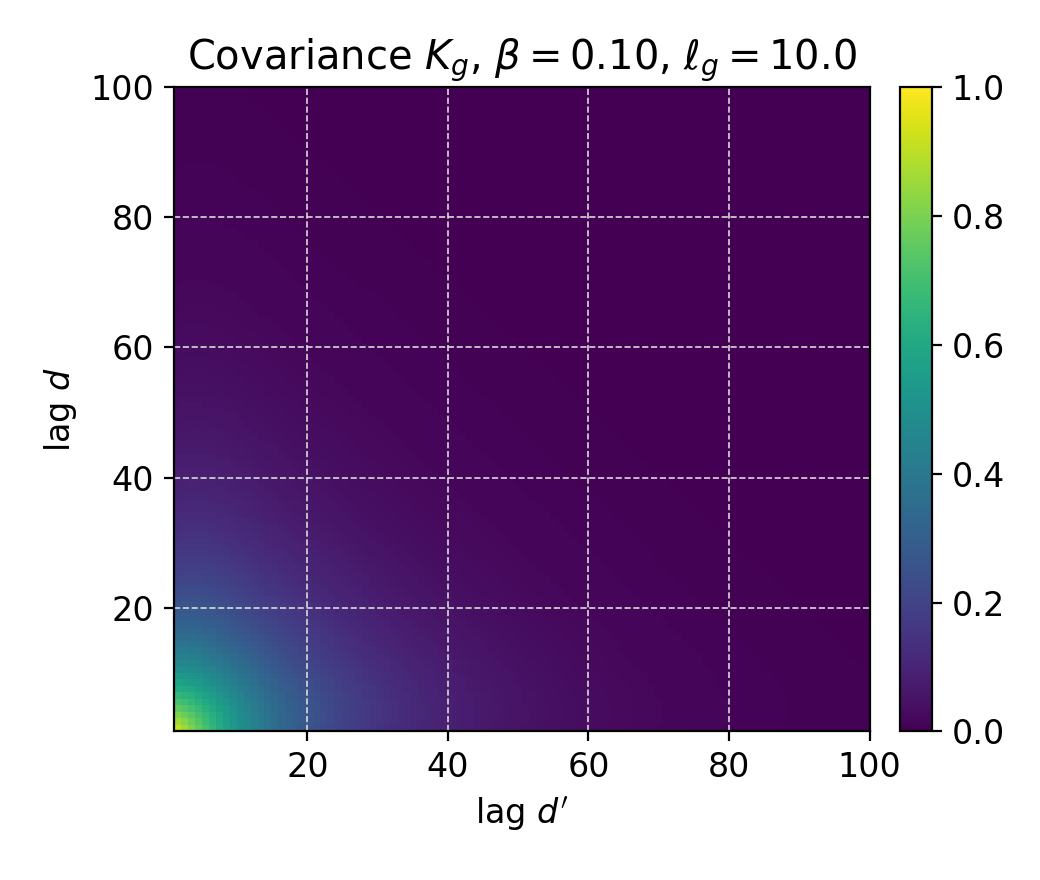}
  \caption{$K_g$ covariance heatmap, $\beta=0.10$, $\ell_g=10$}
\end{subfigure}

\caption{Draws of the GP prior over the lag component $u$: the effect of increasing $\beta$ at fixed $\ell_g$.
Top: draws of $u(d)$; Bottom: corresponding heatmaps for the covariance matrices $K_g$.}
\label{fig:gp-excitation-beta-sweep}
\end{figure}

This appendix reports the hyperparameter search ranges of the forward-validation selection (Table~\ref{tab:hyper-ranges}) and illustrates the GP excitation prior across its warp parameter \(\beta\) (Figure~\ref{fig:gp-excitation-beta-sweep}).

\begin{table}[H]
\centering
\small
\begin{tabular}{p{0.30\linewidth}p{0.64\linewidth}}
\hline
\textbf{Hyperparameter} & \textbf{Search Range} \\
\hline
Baseline level scale \( \sigma_{\mathrm{level}} \) & \{0.5, 1, 2, 5, 10, 20\} \\
Seasonal scale \( \sigma_{\mathrm{season}} \) & \{0.01, 0.05, 0.5, 1, 2, 5, 10, 20\} \\
Linear trend scale \( \sigma_{\mathrm{lin}} \) & \{0, \(10^{-4}\), \(2.5\times10^{-4}\), \(5\times10^{-4}\), \(7.5\times10^{-4}\), \(10^{-3}\), \(2\times10^{-3}\), \(5\times10^{-3}\), \(10^{-2}\)\} \\
NB kernel mass \( K_{\mathrm{NB}} \) & \{0, 0.1, 0.25, 0.5, 0.75, 0.9, 0.98, 1.1, 1.25\} \\
NB kernel mean lag & \{0.25, 0.5, 1, 2, 4, 8, 16, 32, 64\} \\
NB kernel size & \{0.25, 0.5, 1, 2, 5, 10, 25, 50\} \\
Lag attenuation \( \beta \) & \{0.05, 0.1, 0.2, 0.3, 0.5\} \\
GP correction scale \( \sigma_g \) & \{0, 0.25, 0.5, 1, 2, 5, 10\} \\
GP correction lengthscale \( \ell_g \) & \{1, 5, 10, 20, 30\} \\
Softplus link scale \( s \) & \{0.02, 0.05, 0.1, 0.5, 1, 2\} \\
NB2 size \(\kappa\) & \{0.25, 0.5, 1, 2, 5, 10, 20, 50, 100, 200, 500, 1000, 10000, \(10^6\)\} \\
\hline
\end{tabular}
\caption{Hyperparameter ranges for the forward-validation search. The multi-start optimizer searches \emph{continuously} within the box spanned by each row, selecting by forward-validated predictive log-likelihood rather than enumerating a grid; scale parameters are searched on a log scale with strictly positive lower bounds (e.g.\ \(\sigma_{\mathrm{lin}}\in[10^{-8},10^{-2}]\)), so a tabulated \(0\) for a scale denotes the bottom of its box, at which the block is numerically off. The tabulated levels are those an equivalent grid search would enumerate, a comparison on the order of \(10^{8}\) candidate combinations, which the continuous search of Section~\ref*{sec:hyperparameter-selection} avoids. The daily real-data fits additionally select a weekly-baseline scale \(\sigma_{\mathrm{week}}\in[10^{-3},20]\) and, on the two shooting series, the external-covariate block scales, by the same forward-validation search; the synthetic experiments use neither.}
\label{tab:hyper-ranges}
\end{table}

\end{document}